%% file: arxiv.tex
\documentclass[10pt,twocolumn,letterpaper]{article}

\usepackage{iccv}
\usepackage{times}
\usepackage{epsfig}
\usepackage{graphicx}
\usepackage{amsmath}
\usepackage{amssymb}
\usepackage{xpatch}
\usepackage{array}
\usepackage{multirow}
\usepackage{graphicx}
\usepackage{colortbl}
\usepackage{enumitem}

\usepackage{makecell}

\usepackage{capt-of,etoolbox}

\usepackage{overpic}
\usepackage{booktabs}
\usepackage[table]{xcolor}
\definecolor{mypink1}{RGB}{139, 0, 0}

\usepackage[super]{nth}

\definecolor{mycitecolor}{rgb}{0, 0.4, 0.7}
\usepackage[pagebackref=true,breaklinks=true,letterpaper=true,colorlinks,bookmarks=true,citecolor=mycitecolor]{hyperref}

\usepackage[capitalize]{cleveref}
\crefname{section}{Sec.}{Secs.}
\Crefname{section}{Section}{Sections}
\Crefname{table}{Table}{Tables}
\crefname{table}{Tab.}{Tabs.}

\definecolor{ncar}{RGB}{255, 158, 0}
\definecolor{ntruck}{RGB}{255, 99, 71}
\definecolor{ntrailer}{RGB}{255, 140, 0}
\definecolor{nbus}{RGB}{255, 69, 0}
\definecolor{nconstruct}{RGB}{233, 150, 70}
\definecolor{nbicycle}{RGB}{220, 20, 60}
\definecolor{nmotor}{RGB}{255, 61, 99}
\definecolor{npedestrian}{RGB}{0, 0, 230}
\definecolor{ntraffic}{RGB}{47, 79, 79}
\definecolor{nbarrier}{RGB}{112, 128, 144}
\definecolor{ndriveable}{RGB}{0, 207, 191}
\definecolor{nother}{RGB}{175, 0, 75}
\definecolor{nsidewalk}{RGB}{75, 0, 75}
\definecolor{nterrain}{RGB}{112, 180, 60}
\definecolor{nmanmade}{RGB}{222, 184, 135}
\definecolor{nvegetation}{RGB}{0, 175, 0}

\newcommand{\colortable}{\cellcolor[RGB]{220,220,220}}
\iccvfinalcopy 

\newcommand{\hy}[1]{\textcolor{blue}{[HY]: #1}}

\def\iccvPaperID{1754} 

\ificcvfinal\fi


\graphicspath{{figs/}}
\begin{document}

\title{Scene as Occupancy}

\author{
Chonghao Sima$^{1,3\ast\dagger}$,
Wenwen Tong$^{2\ast}$,
Tai Wang$^{1,4}$,
Li Chen$^{1,3}$, 
Silei Wu$^{2}$, \\
Hanming Deng$^2$,
Yi Gu$^{1}$,
Lewei Lu$^{2}$,
Ping Luo$^{3}$,
Dahua Lin$^{1,4}$,
Hongyang Li$^{1\dagger}$ \\
[2mm]
$^1$~Shanghai AI Laboratory \quad 
$^2$~SenseTime Research \\
$^3$~The University of Hong Kong \quad
$^4$~The Chinese University of Hong Kong \\
\normalsize{
$^\ast$Equal contribution \quad $^\dagger$Project lead}
\\
\normalsize{\url{https://github.com/OpenDriveLab/OccNet}
}
}

\maketitle

\ificcvfinal\thispagestyle{empty}\fi

\begin{abstract}
\input{sections/abstract_smch}
\end{abstract}

\section{Introduction}

\begin{figure}[!th]
  \centering
  \includegraphics[width=.95\linewidth]{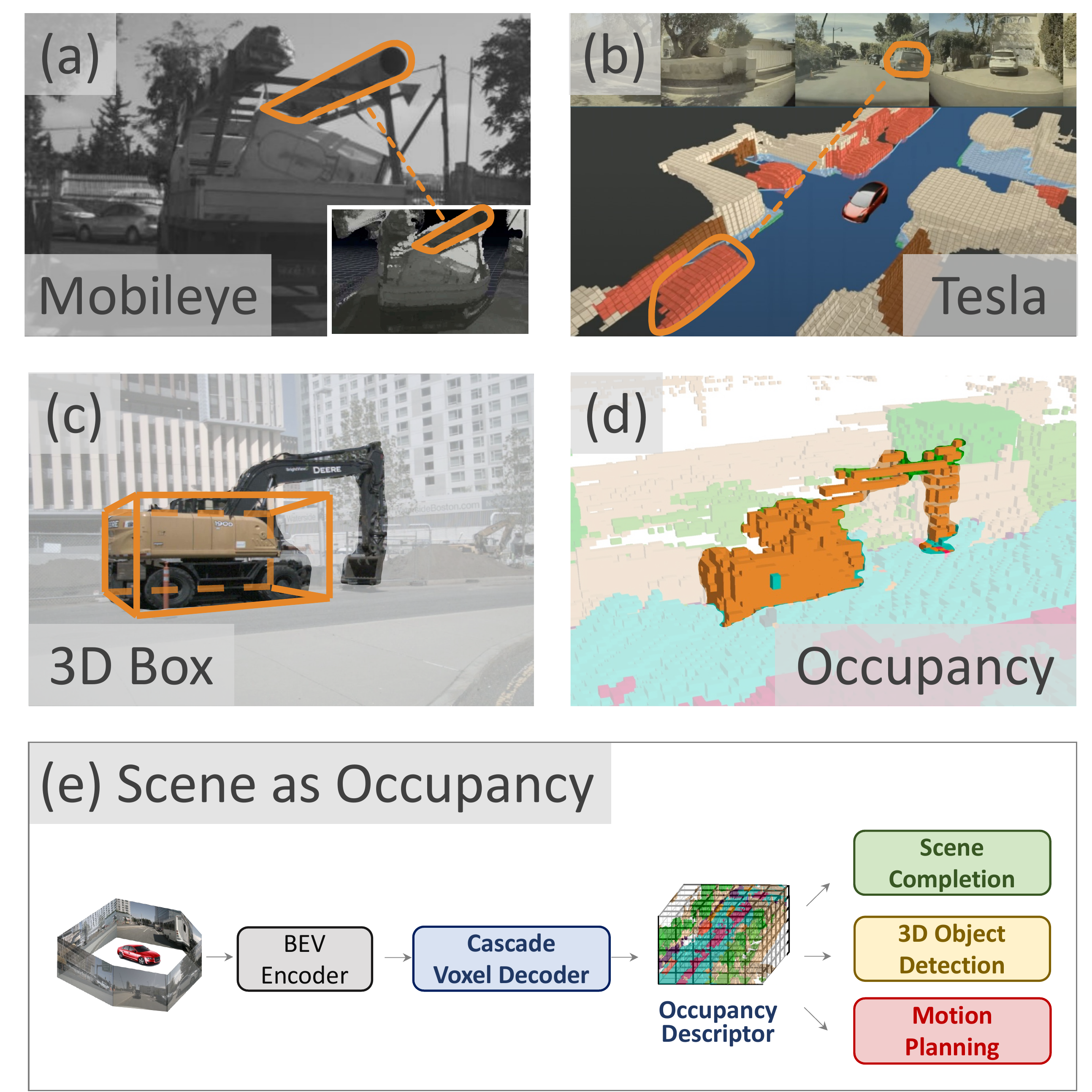}
  \caption{
\textbf{Scene as Occupancy.}
Representing objects as ViDAR \textbf{(a)}  or 3D occupancy \textbf{(b)}  has been endorsed by industry \cite{mobile2020ces,tesla_ai_day}, 
due to the fact that conventional 3D bounding box cannot describe in detail irregular vehicles in daily driving scenes, \textit{e.g.}, protruding tail in \textbf{(a)} or (\textbf{c)}.
Defining the 3D world as \textit{Occupancy} in \textbf{(d)}  serves better to represent obstacles and avoid collision. 
%
In this paper, we envision \textit{Occupancy} as a general \textit{Scene Descriptor} as in \textbf{(e)}  for a wide span of driving tasks \textit{beyond} detection, such as planning,
and witness performance gain compared to previous 
alternatives.
  }
  \label{fig:motivation}
\end{figure}

When you are driving on the road, how would you describe the scene in 3D space through your eyes?
Human driver can easily describe the environment by ``There is a Benz on the left side of my car in around 5 inches", ``There is a truck carrying huge protruding gas pipe on the rear, in around 50 meters ahead'' and so on.
Having the ability to describe the real world in a ``There is'' form is essential for making safe autonomous driving (AD) a reality.
This is non-trivial for vision-centric AD systems due to the diverse range of entities present in the \emph{Scene}, including vehicles such as cars, SUVs, and construction trucks, as well as static barriers, pedestrians, background buildings and vegetation.
Quantizing the 3D scene into structured cells with semantic labels attached, termed as \emph{3D Occupancy}, is an intuitive solution, and this form is also advocated in the industry communities such as Mobileye~\cite{mobile2020ces} and Tesla~\cite{tesla_ai_day} . 
Compared to the 3D box that oversimplifies the shape of objects, 3D occupancy is geometry-aware, depicting different objects and background shapes via the 3D cube collections with different geometric structure.
As illustrated in Figure~\ref{fig:motivation}(c-d), 3D box can only describe the main body of the construction vehicle, while 3D occupancy can preserve the detail of its crane arm.
%
Other conventional alternatives, such as point cloud segmentation and bird's-eye-view (BEV) segmentation, while being widely deployed in the context of AD, have their limitations in cost and granularity, respectively.
A detailed comparison can be referred in Table~\ref{tab:compare_to_other_form}.
%
Such evident advantages of 3D occupancy encourage an investigation into its potential for augmenting conventional perception tasks and downstream planning.

Similar works have discussed 3D occupancy at an initial stage.
Occupancy grid map, a similar concept in Robotics, is a typical representation in mobile navigation~\cite{moras2015grid} but only serves as the search space of planning.
3D semantic scene completion (SSC) 
\cite{song2016ssc}
can be regarded as a perception task to evaluate the idea of 3D occupancy.
Exploiting temporal information as geometric prior is intuitive for the vision-centric models to reconstruct the geometry-aware 3D occupancy, yet previous attempts~\cite{huang2023tri,li2023voxformer,cao2022monoscene,miao2023occdepth} have failed to address this.
A coarse-to-fine approach is also favorable in improving 3D geometric representation at affordable cost, 
while it is ignored by one-stage methods~\cite{huang2023tri,miao2023occdepth,cao2022monoscene}.
In addition, the community still seeks a practical approach to evaluate 3D occupancy in a full-stack autonomous driving spirit as vision-centric solutions~\cite{yang2022goal} prevail.

Towards these issues aforementioned, we propose \emph{OccNet}, a multi-view vision-centric pipeline with a cascade voxel decoder to reconstruct 3D occupancy with the aid of temporal clues, and task-specific heads supporting a wide range of driving tasks.
The core of \emph{OccNet} is a compact and representative 3D occupancy embedding to describe the 3D scene.
To achieve this, unlike straightforward voxel feature generation from image features or sole use of BEV feature as in previous literature~\cite{li2022bevformer,chen2022persformer,wang2022monocular}, \emph{OccNet} employs a cascade fashion to decode 3D occupancy feature from BEV feature.
The decoder adopts a progressive scheme to recover the height information with voxel-based temporal self-attention and spatial cross-attention, bundled alongside a deformable 3D attention module for efficiency.
Equipped with such a 3D occupancy descriptor, \emph{OccNet} simultaneously supports general 3D perception tasks and facilitates downstream planning task, \textit{i.e.}, 3D occupancy prediction, 3D detection, BEV segmentation, and motion planning.
%
%
For fair comparison across methods,
we build \emph{OpenOcc}, a 3D occupancy benchmark with dense and high-quality annotations, based on nuScenes dataset~\cite{caesar2020nuscenes,fong2022panoptic}.
It comprises 34149 annotated frames with over 1.4 billion 3D occupancy cells, each assigned to one of 16 classes to describe foreground objects and background stuff.
Such dense and semantic-rich annotations leverage vision models towards superior 3D geometry learning, compared to the sparse alternative.
It takes object motion into consideration with directional flow annotations as well, being extensible to the planning task.
%
%

We evaluate OccNet on OpenOcc benchmark, and empirical studies demonstrate the superiority of 3D occupancy as a scene representation over traditional alternatives from three aspects:
1) \emph{Better perception.}
3D occupancy facilitates the acquisition of 3D geometry from vision-only models, as evidenced by the point cloud segmentation performance comparable with LiDAR-based methods and the enhanced 3D detection performance with occupancy-based pre-training or joint-training.
2) \emph{Better Planning.}
More accurate perception also translates into improved planning performance.
3) \emph{Dense is better.}
Dense 3D occupancy proves more effective than sparse form in supervising vision-only models. 
%
On the OpenOcc benchmark, OccNet outperforms state-of-the-art, \textit{e.g.} TPVFormer~\cite{huang2023tri}, with a relative improvement of 14\% in the semantic scene completion task.
Compared with FCOS3D~\cite{wang2021fcos3d}, the detection model performance pre-trained on OccNet increases by about 10 points when fine-tuned on small-scale data.
For the motion planning task based on 3D occupancy, we can reduce the collision rate by 15\%-58\% compared with the planning policy based on BEV segmentation or 3D boxes.

\iftrue
To sum up, our contributions are two folds: 
%
%
\textbf{(1)} 
We propose OccNet, a vision-centric pipeline with a cascade voxel decoder to generate 3D occupancy using temporal clues. 
It  better captures the fine-grained details of the physical world and supports a wide range of driving tasks.
\textbf{(2)} 
Based on the proposed OpenOcc benchmark with dense and high-quality annotations, we demonstrate the effectiveness of OccNet with an evident performance gain upon perception and planning tasks.
An initial conclusion is that 3D occupancy, as scene representation, is superior to conventional alternatives.
%
\fi

\begin{table}[tb]
\small
\centering
\setlength{\tabcolsep}{2.5pt}
\setlength{\aboverulesep}{0pt} 
\setlength{\belowrulesep}{0pt} 
\resizebox{\linewidth}{!}{
\begin{tabular}{l|c|c|c|c|c}
    \toprule
         {Representation}   & \makecell{Output \\ space} & \makecell{Foreground \\ object} & \makecell{Background \\ \& Mapping} & \makecell{Description \\ Granularity}  & \makecell{ Require point \\ cloud input} \\
         \midrule
         3D Box  & 3D & \checkmark & - &  $0.4 \sim 12m$ & - \\
         BEV Seg.  & BEV & \checkmark & \checkmark & $0.5m \sim 1m$  & - \\
         Point cloud  & 3D & \checkmark & - & $\sim 0.02m$  & \checkmark \\
         \midrule
         \rowcolor[RGB]{220,220,220}{3D Occupancy}& 3D & \checkmark & \checkmark & $0.25 \sim 0.5m$  & - \\
        \bottomrule
	\end{tabular}
 }
\vspace{0.5mm}
	\caption{\textbf{Comparison on different representations.} 3D Occupancy unifies foreground objects and background stuff into a fine-grain and dense voxel space, and is input-modality-agnostic.}
	\label{tab:compare_to_other_form}
\end{table}

\begin{figure*}[th]
  \centering
  \includegraphics[width=\linewidth]{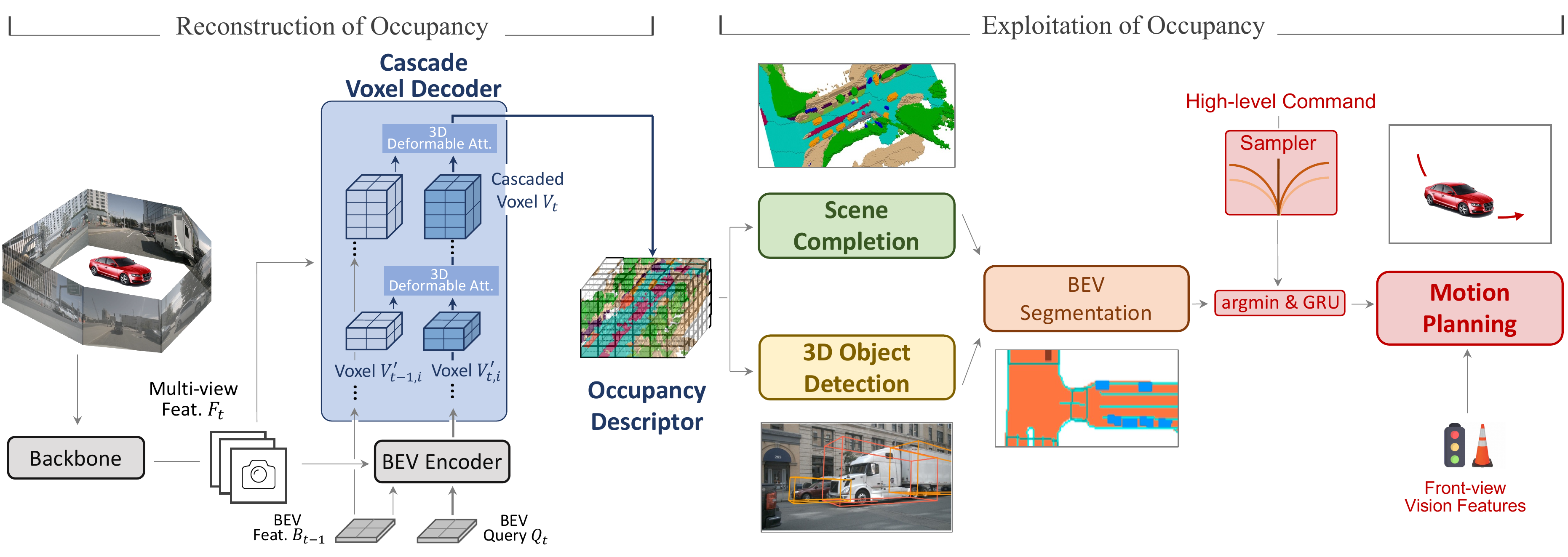}
  \caption{\textbf{OccNet pipeline.} 
  The core of OccNet is to obtain a representative Occupancy Descriptor and apply it for various driving tasks.
  Our proposed algorithm consists of two stages. \textbf{I. Reconstruction of Occupancy.} 
Given multiple visual inputs, we first generate features from the BEV encoder. Voxel Decoder is performed in a cascade fashion where voxels are refined progressively. A 3D deformable attention (att.) unit serves similar functionality as does in 2D case. Temporal voxels $V_{t-1}$ are also incorporated. Some connections are omitted for brevity. See context for details.
\textbf{II. Exploitation of Occupancy.} 
Equipped with the occupancy descriptor, we can proceed tasks including semantic scene completion and 3D object detection. Compacting them in BEV space would obtain a BEV segmentation map, which can be directly fed into the planning pipeline \cite{st-p32022}. Such a design can ensure desirable improvement in planning task.
  }
  \label{fig:framework}
\end{figure*}

\section{Related Work}
%
%

%

\noindent\textbf{3D object detection} 
\cite{Shi2020PVRCNNPF,wang2021fcos3d,li2022bevformer,liu2022bevfusion} adopts 
3D boxes as the objective of perception in AD 
since the box-form is well structured for downstream rule-based approaches.
Such a representation abstracts 3D objects with different shapes into standardized cuboids, hence only cares about foreground objects and oversimplifies object shape.
In contrast, 3D occupancy is a fine-grained description of the physical world 
and can differentiate objects with various shapes. 
%

\noindent\textbf{LiDAR segmentation} 
\cite{zhu2021cylindrical,milioto2019rangenet++} is tasked as point-level 3D scene understanding.
%
It requires point cloud as input, which is expensive and less portable.
Since LiDAR inherently suffers from limited sensing range and sparsity in 3D scene description, it is not friendly to holistic 3D scene semantic understanding~\cite{song2016ssc} using such a pipeline.
%

%
%

\noindent\textbf{3D reconstruction and rendering.}
Inferring the 3D geometry of objects or scenes from 2D images \cite{Hartley2004,mildenhall2020nerf} is prevailing yet challenging for many years
in computer vision.
%
Most approaches in this domain~\cite{mueller2022instant,cao2022scenerf,Tancik2022blocknerf} cope with a \textit{single} object or scene. For AD application, 
this is not feasible since it requires strong generalization ability.
Note that 3D reconstruction and rendering concentrates more on the quality of the scene geometry and visual appearance. It pays less attention to model efficiency and semantic understanding.
%
%

\noindent\textbf{Semantic Scene Completion.} The definition of occupancy prediction discussed in this work shares the most resemblance with SSC \cite{song2016ssc}.
MonoScene~\cite{cao2022monoscene} first adopts U-Net to infer from a single monocular RGB image the dense 3D occupancy with semantic labels. There is a burst of related works released in arXiv recently. We deem them as concurrent and briefly discuss below.
%
VoxFormer \cite{li2023voxformer} utilizes the depth estimation to set voxel queries in a two-stage framework.
OccDepth~\cite{miao2023occdepth} also adopts a depth-aware spirit in a stereo setting with distillation to predict semantic occupancy.
TPVFormer~\cite{huang2023tri} employs LiDAR-based sparse 3D occupancy as the supervision and proposes a tri-perspective view representation to obtain features.
Wang \textit{et al.} \cite{wang2023openoccupancy} provides a well human-crafted occupancy benchmark that could facilitate the community.
%
%
%
%

Despite different settings from ours with work conducted on
Semantic-KITTI~\cite{behley2019semantickitti} and NYUv2~\cite{Silberman2012nyuv2} 
(monocular or RGB-D), 
%
prior or concurrent literature unanimously neglect the adoption of temporal context. Utilizing history voxel features is straightforward; it is verified by Tesla \cite{tesla_ai_day}. Yet there is no technical details or report to the public. Moreover, we position our work to be the first to investigate occupancy as a general descriptor that could enhance multiple tasks {\textit{beyond}} detection.

%
%

\section{Methodology}
In this paper, we propose an effective and general framework, named OccNet, which obtains robust occupancy features from images and supports multiple driving tasks, as shown in Figure~\ref{fig:framework}.
Our method comprises two stages, \textit{Reconstruction of Occupancy} and \textit{Exploitation of Occupancy}.
We term the bridging part as \textit{Occupancy Descriptor}, a unified description of the driving scene.
%

\noindent\textbf{Reconstruction of Occupancy.}
The goal of this stage is to obtain a representative occupancy descriptor for supporting downstream tasks.
Motivated by the fast development in BEV perception~\cite{li2022bevformer,chen2022persformer,liang2022bevfusion}, OccNet is designed to exploit that gain for the voxel-wise prediction task in 3D space.
To achieve this, the sole usage of BEV feature in downstream tasks, as the simplest architecture, is not suitable for height-aware task in 3D space. 
Going from one extreme to another, directly constructing voxel feature from images has huge computational cost.
We term these two extreme as \emph{BEVNet} and \emph{VoxelNet}, and
the design of OccNet finds a balance between them, achieving the best performance with affordable cost.
The reconstruction stage first extracts multi-view feature $F_t$ from surrounding images, and feeds them into BEV encoder along with history BEV feature $B_{t-1}$ and current BEV query $Q_t$ to get current BEV feature.
The BEV encoder follows the structure of BEVFormer \cite{li2022bevformer}, where history BEV feature $B_{t-1}$, current BEV query $Q_t$ and image feature $F_t$ go through a spatial-temporal-transformer block to get current BEV feature.
Then, the image feature, the history and current BEV feature are together decoded into occupancy descriptor via \textit{Cascade Voxel Decoder}.
Details of the decoder is presented in Sec.~\ref{sec:cascade_voxel_decoder}.

\noindent\textbf{Exploitation of Occupancy.}
A wide range of driving tasks can be deployed based on the reconstructed occupancy descriptor.
%
%
%
Inspired by Uni-AD~\cite{yang2022goal}, an explicit design of each representation is preferred.
Intuitively, 3D semantic scene completion~\cite{song2016ssc} and 3D object detection are attached upon the occupancy descriptor.
Squeezing 3D occupancy grid map and 3D boxes along the height generates a BEV segmentation map.
Such a map can be directly fed into motion planning head, along with sampler of high-level command, resulting in the ego-vehicle trajectory via \textit{argmin} and GRU module.
 Detailed illustration 
 is provided 
 in Sec.~\ref{sec:exploitation_of_occupancy}.

\subsection{Cascade Voxel Decoder}\label{sec:cascade_voxel_decoder}
To obtain a better voxel feature effectively and efficiently, we design a cascade structure in the decoder to progressively recover the height information in voxel feature. 

\noindent\textbf{From BEV to Cascaded Voxel}.
Based on the observation that directly using BEV feature or directly reconstructing voxel feature from perspective view suffers from performance or efficiency drop (see our ablation in Table~\ref{tab:irregular}), we break this reconstruction from BEV feature ($B_t \in \mathbb{R}^{H \times W \times C_{\text{BEV}}}$) to the desired voxel feature ($V_t \in \mathbb{R}^{Z \times H \times W \times C_{\text{Voxel}}}$) into $N$ steps, named a cascade structure.
Here $H$ and $W$ are the 2D spatial shape of BEV space, $C$ the feature dimension and $Z$ the desired height of voxel space.
Between the input BEV feature and the desired cascaded voxel feature, we term the intermediate voxel feature with different height as $V_{t,i}^{'} \in \mathbb{R}^{Z_i \times H \times W \times C_i}$, where $Z_i$ and $C_i$ are uniformly distributed between $\{1, N\}$ and $\{C_{\text{BEV}}, C_{\text{Voxel}}\}$ respectively.
%
%
%
%
As shown in Figure~\ref{fig:framework}, the $B_{t-1}$ and $B_t$ are lifted into $V_{t-1,i}^{'}$ and $V_{t,i}^{'}$ via feed-forward network, go through the i-$th$ voxel decoder to obtain a refined $V_{t,i}^{'}$, and the later steps follow the same scheme.
Each voxel decoder comprises voxel-based temporal self-attention and voxel-based spatial cross-attention modules, and refines $V_{t,i}^{'}$ with history $V_{t-1,i}^{'}$ and image feature $F_t$ respectively.
Step by step, the model gradually increases $Z_i$ and decreases $C_i$ to learn the final occupancy descriptor $V_t$ effectively and efficiently.

\noindent\textbf{Voxel based Temporal Self-Attention.}
The temporal information is crucial to represent the driving scene accurately \cite{li2022bevformer}. 
Given the history voxel feature $V_{t-1,i}^{'}$, we align it to the current occupancy features $V_{t,i}^{'}$ via the position of ego-vehicle. 
For a typical self-attention, each query attends to every key and value, so the computation cost is very huge and even increases $Z^2$ times in 3D space compared to the 2D case.
To alleviate the computation cost, we design a voxel-based efficient attention, termed as \textit{3D Deformable Attention} (3D-DA in short), to handle the computational burden.
By applying it in the voxel-based temporal self-attention, we ensure that
each voxel query only needs to interact with local voxels of interest, making the computational cost affordable.

\noindent\textbf{{3D Deformable Attention.}}
%
We extend the traditional 2D deformable attention \cite{Zhu2021DeformableDD} to 3D form.
%
Given a voxel feature $V_{t,i}^{'} \in \mathbb{R}^{Z_i \times H \times W \times C_i} $, a voxel query with feature $\boldsymbol{q} \in \mathbb{R}^C_i$ and 3D referent point $\boldsymbol{p}$, the 3D deformable attention is represented by:
\begin{equation}
    \operatorname{3D-DA}(\boldsymbol{q}, \boldsymbol{p}, V_{t,i}^{'})
    = \sum_{m=1}^{M} W_m \sum_{k=1}^{K} A_{mk} W_k^{'} V_{t,i}^{'}(\boldsymbol{p}+\Delta\boldsymbol{p}_{mk}),
\end{equation}
where $M$ is the number of attention heads, $K$ is the sampled key number 
with $K \ll Z_iHW$, $W_m \in \mathbb{R}^{C_i\times (C_i/M)}$ and  $W_k \in \mathbb{R}^{(C_i/M) \times C_i}$ are the learning weights, 
$A_{mk}$ is the normalized attention weight, and 
$\boldsymbol{p}+\Delta \boldsymbol{p}_{mk}$ is the learnable sample point position in 3D space, in which the feature is computed by trilinear interpolation from the voxel feature.

\noindent\textbf{Voxel-based Spatial Cross-Attention.}
In the cross attention, the voxel feature $V_{t,i}^{'}$ interacts with the multi-scale image features $F_t$ with 2D deformable attention~\cite{Zhu2021DeformableDD}.
%
%
Each i-$th$ decoder directly samples $N_{ref,i}$ 3D points from the corresponding voxel to the image view, and interact with the sampled image feature.
Such a design maintains the height information and ensures the learning of voxel-wise feature.



\subsection{Exploiting Occupancy on Various Tasks}\label{sec:exploitation_of_occupancy}
The OccNet depicts the scene in 3D space with fine-grained occupancy descriptor, which can be fed into various driving tasks without excessive computational overhead.

\noindent\textbf{Semantic Scene Completion.}
For simplicity, we design the MLP head to predict the semantic label of each voxel, and apply the Focal loss~\cite{Lin2017FocalLF} to balance the huge numerical inequality between occupied and empty voxels. 
In addition, the flow head with $L_1$ loss are attached to estimate the flow velocity per occupied voxels.

\noindent\textbf{3D Object Detection.}
Inspired by the head design in BEVFormer~\cite{li2022bevformer}, we compact the occupancy descriptor into BEV, then apply a query-based detection head (an invariant of Deformable DETR~\cite{Zhu2021DeformableDD}) to predict the 3D boxes.

\noindent\textbf{BEV segmentation.}
Following the spatial-temporal-fusion perception structure in ST-P3~\cite{st-p32022}, map representation and semantic segmentation are predicted from the BEV feature as in 3D object detection.
The BEV segmentation head includes the drivable-area head and the lane head for map representation, the vehicle segmentation head and the pedestrian segmentation head for semantic segmentation.

\noindent\textbf{Motion Planning.}
For motion planning task, either the predicted occupancy results in SSC or 3D bounding box can be transformed into the BEV segmentation, as shown in~\ref{fig:framework}.
The 3D occupancy results is squeezed along the height dimension and the 3D boxes as well.
All the semantic labels per BEV cell from either 3D occupancy or 3D boxes are turned into a 0-1 format, where 1 indicates the cell is occupied and 0 for empty.
Then, such a BEV segmentation map is applied to the safety cost function, and we compute the safety, comfort and progress cost on the sampled trajectories.
Note that compared to 3D boxes, the richer background information in occupancy scene completion leads to the more comprehensive safety cost function, and thus the safety cost value is needed to be normalized between these two kinds of BEV segmentation. 
All candidate trajectories are sampled by random velocity, acceleration, and curvature. Under the guidance of high-level command including forward, turn left and turn right, the trajectory corresponding to the specific command with the lowest cost will be output.
GRU refinement enabled with the front-view vision feature is further performed on this trajectory as ST-P3~\cite{st-p32022} to obtain the final trajectory.


\section{OpenOcc: 3D Occupancy Benchmark}
To fairly evaluate the performance of occupancy across literature, we introduce the {first} 
3D occupancy benchmark named as \textit{OpenOcc} built on top of the prevailing nuScenes dataset~\cite{caesar2020nuscenes,fong2022panoptic}.
Compared with existing counterparts such as SemanticKITTI~\cite{behley2019semantickitti} with only front camera,  OpenOcc provides surrounding camera views with the corresponding 3D occupancy and flow annotations.

\subsection{Benchmark Overview}
\begin{figure}
  \centering
  \includegraphics[width=0.9\linewidth]{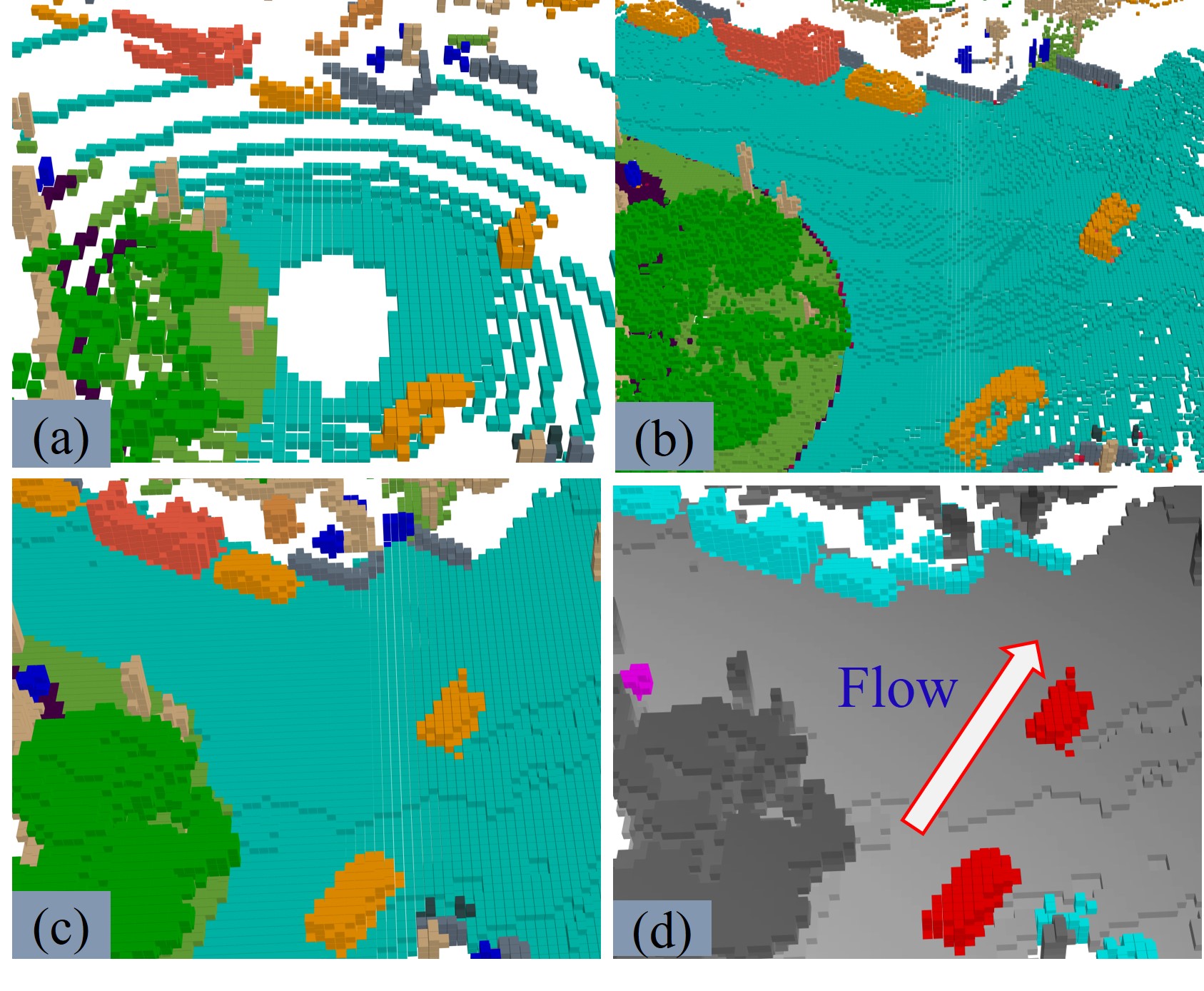}
  \caption{\textbf{Visual comparison on 3D occupancy annotations.} Compared to (a) sparse occupancy~\cite{huang2023tri} and (b) OccData~\cite{Fang2023}, we generate (c) dense and high-quality 
  annotations with (d) the additional flow annotation of foreground objects, which can be applied for motion planning.
  }
  \label{fig:compare_gt}
\end{figure}

We generate occupancy data with dense and high quality occupancy annotations utilizing the sparse LiDAR information and 3D boxes. It comprises 34149 annotated frames for all 700 training and 150 validation scenes. We annotate over 1.4 billion voxels and 16 classes in the benchmark, including 10 foreground objects and 6 background stuffs. Moreover, we take the foreground object motion into consideration with additional flow annotation of object voxels.
We compare our occupancy data with other benchmark in Table~\ref{tab:compare_gt}, indicating that our benchmark can provide the most complete representation of the scene including the occupancy and flow information.
As depicted in Figure~\ref{fig:compare_gt}, 
SparseOcc \cite{huang2023tri} only utilized the sparse key frame LiDAR data to voxelize the 3D space, which is too sparse to represent the 3D scene.
In comparison, our occupancy can represent the complete scene with flow information and capture the local fine grained scene geometry with high quality.

\begin{table}[t]
  \centering    
  \resizebox{\linewidth}{!}{
    \begin{tabular}{l|cccc}
    \toprule
    Dataset &  Multi-view  & Scenes & Flow  & Density\\
    \midrule  
    SemanticKITTI \cite{behley2019semantickitti} & - & 22 & - & - \\
    SparseOcc~\cite{huang2023tri} & \checkmark & 850 & - & $\sim 0.11$\\
    OccData~\cite{Fang2023} &\checkmark & 850 & -&  $\sim 0.76$\\
    \rowcolor[RGB]{220,220,220}OpenOcc (Ours)  & \checkmark & 850 & \checkmark &  1\\
    \bottomrule
  \end{tabular}
  }
  \vspace{0.5mm}
  \caption{\textbf{Comparison of OpenOcc with existing benchmarks.} 
  \emph{Multi-view} denotes the dataset that use muti-view image as input.
  \emph{Flow} represent the flow annotation is given in the dataset.
  The density measures the voxel density in the dataset.
  }
  \label{tab:compare_gt}
\end{table}

\begin{table*}[t]
\small
\setlength{\tabcolsep}{3.5pt}
\centering
\resizebox{\textwidth}{!}{

\begin{tabular}{@{}l|c|c|>{\colortable}c|cccccccccccccccc@{}}
\toprule
Method                  & Backbone  & \text{IoU}$_{geo}$   & mIoU  &\rotatebox{90}{barrier} & \rotatebox{90}{bicycle} &  \rotatebox{90}{bus}   & \rotatebox{90}{car}   & \rotatebox{90}{const. veh.} & \rotatebox{90}{motorcycle} & \rotatebox{90}{pedestrian} & \rotatebox{90}{traffic cone} & \rotatebox{90}{trailer} & \rotatebox{90}{truck} & \rotatebox{90}{driv. surf.} & \rotatebox{90}{other flat} & \rotatebox{90}{sidewalk} & \rotatebox{90}{terrain} & \rotatebox{90}{manmade} & \rotatebox{90}{vegetation} \\
\midrule
BEVDet4D \cite{huang2022bevdet4d}                 & ResNet50  & 18.27 & 9.85  & 13.56   & 0.00       & 13.04 & 26.98 & 0.61                  & 1.20        & 6.76       & 0.93          & 1.93    & 12.63 & 27.23              & 11.09       & 13.64    & 12.04   & 6.42    & 9.56       \\
BEVDepth \cite{li2022bevdepth}               & ResNet50  & 23.45 & 11.88 & 15.15   & 0.02    & 20.75 & 27.05 & 1.10                   & 2.01       & 9.69       & 1.45          & 1.91    & 14.31 & 31.92              & 7.88        & 17.08    & 16.27   & 8.76    & 14.75      \\
BEVDet \cite{huang2021bevdet}                & ResNet50  & 27.46 & 12.49 & 16.06   & 0.11    & 18.27 & 21.09 & 2.62                  & 1.42       & 7.78       & 1.08          & 3.4     & 13.76 & 33.89              & 10.84       & 17.55    & 22.03   & 11.72   & 18.15      \\
\rowcolor[RGB]{220,220,220}OccNet (ours)  & ResNet50 & \textbf{37.69} & \textbf{19.48} & \textbf{20.63}   & \textbf{5.52}   & \textbf{24.16} & \textbf{27.72} & \textbf{9.79}                 & \textbf{7.73}      & \textbf{13.38}      & \textbf{7.18}         & \textbf{10.68}   & \textbf{18.00} &  \textbf{46.13}              & \textbf{20.6}        & \textbf{26.75}    & \textbf{29.37}   & \textbf{16.90}    & \textbf{27.21} \\
\midrule
TPVFormer$^{ \ast }$~\cite{huang2023tri}              & ResNet101 & 37.47 & 23.67 & 27.95   & 12.75   & 33.24 & \textbf{38.70}  & 12.41                 & 17.84      & 11.65      & 8.49          & \textbf{16.42}   & \textbf{26.47} & 47.88              & 25.43       & 30.62    & 30.18   & 15.51   & 23.12      \\
\rowcolor[RGB]{220,220,220}OccNet (ours)  & ResNet101 & \textbf{41.08} &\textbf{26.98} & \textbf{29.77}   & \textbf{16.89}   & \textbf{34.16} & 37.35 & \textbf{15.58}                 & \textbf{21.92}      & \textbf{21.29}      & \textbf{16.75}         & 16.37   & {26.23} & \textbf{50.74}              & \textbf{27.93}       & \textbf{31.98}    & \textbf{33.24}   & \textbf{20.80}    & \textbf{30.68}     \\
\bottomrule
\end{tabular}
}
\vspace{0.5mm}
\caption{\textbf{3D Occupancy Prediction in terms of Semantic Scene Completion.} The semantic occupancy prediction and geometric prediction metrics are compared for models with RGB input. OccNet significantly outperforms previous SOTAs in terms of mIoU and $\text{IoU}_{geo}$. Methods with $^{ \ast }$ stands for training and evaluating on OpenOcc dataset.
}
\label{tab:occ_method}
\end{table*}

\begin{figure*}
  \centering
    \includegraphics[width=0.95\linewidth]{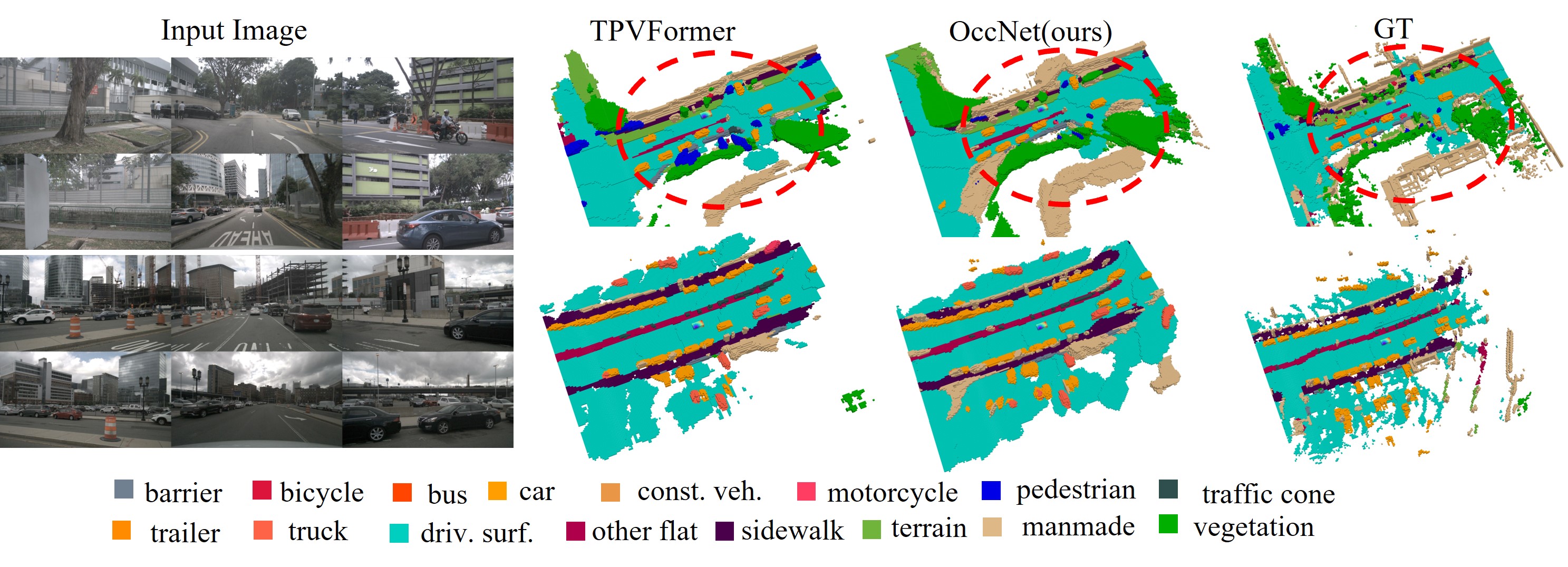}
  \caption{\textbf{Qualitative results of occupancy prediction.} Our method outperforms TPVFormer \cite{huang2023tri} in terms of scene details and the semantic classification accuracy of foreground objects, such as the pedestrian in the dashed region.
  }
  \label{fig:compare_occ_method}
\end{figure*}

\subsection{Generating High-quality Annotation}
\noindent\textbf{Independent Accumulation of Background and Foreground.} 
To generate dense representation, it is intuitive to accumulate all sparse LiDAR points from the key frame and intermediate frame to obtain the dense representation \cite{behley2019semantickitti}.
However, directly accumulating points from intermediate frame by coordinate transformation is problematic owing to the existence of moving objects.
We propose to split the LiDAR point into the static background points and foreground points based on 3D box
and accumulate them separately. 
Then we can accumulate static background points in the global world system and object points in the object coordinate system to generate dense points.

\noindent\textbf{Generation of Annotation.} 
Given dense background and object points, we first voxelize the 3D space and label the voxel based on the majority vote of labelled points in the voxel.
Different with existing benchmark with only occupancy labels, we annotate the flow velocity of voxel based on the 3D box velocity to faciliatate the downstream task such as motion planning.
Only Using key frame will cause sparsity of generated occupancy data, thus 
we annotate the voxel with unlabeled LiDAR points from intermediate frame based on the surrounding labelled voxels to further improve the data density.
In addition, as nuScenes has the issue of missing translation in the z-axis, we refine the occupancy data by completing the scene, such as filling the holes on the road for higher quality.
Moreover, we set part of voxels as invisible from the camera view by tracing the ray, which is more applicable for the task with camera input.

\section{Experiments}

\noindent\textbf{Benchmark Details.} We select a volume of $\mathcal{V}=[-50\text{m}, 50\text{m}] \times [-50\text{m}, 50\text{m}] \times [-5\text{m}, 3\text{m}]$ in LiDAR coordinate system for occupancy data generation, and voxelize the 3D space by the resolution of $\Delta s=0.5\text{m}$ into $200 \times 200 \times 16$ voxels to represent the 3D space. Evaluation metric can be referred in Supplementary.

\noindent\textbf{OccNet Details.} Following the experimental setting of BEVFormer \cite{li2022bevformer}, we use two types of backbone: ResNet50 \cite{he2016resnet} initialized from ImageNet \cite{deng2009imagenet}, and ResNet101-DCN \cite{he2016resnet} initialized from FCOS3D \cite{wang2021fcos3d}.
We define the BEV feature as $B_t$ with $H=200$, $W=200$, and $C_{\text{BEV}}=256$.
For the decoder, we design $N=4$ occupancy feature maps $V_{t,i}^{'} \in \mathbb{R}^{Z_i \times H \times W \times C_i} $ with $Z_i=2^{i}$, $C_1=C_2=128$, $C_3=C_4=64$.
For the voxel-spatial cross attention, we sample $N_{ref,i}=4$ points in each queried voxel.
By default, we train OccNet with 24 epochs with a learning rate of $2\times 10^{-4}$.


\subsection{Main Results}

\begin{table*}[t]
\small
\setlength{\tabcolsep}{3.5pt}
\centering
\resizebox{\textwidth}{!}{%
\begin{tabular}{@{}l|c|c|cccccccccccccccc@{}}
\toprule
Method                        & Input & mIOU  &\rotatebox{90}{barrier} & \rotatebox{90}{bicycle} &  \rotatebox{90}{bus}   & \rotatebox{90}{car}   & \rotatebox{90}{const. veh.} & \rotatebox{90}{motorcycle} & \rotatebox{90}{pedestrian} & \rotatebox{90}{traffic cone} & \rotatebox{90}{trailer} & \rotatebox{90}{truck} & \rotatebox{90}{driv. surf.} & \rotatebox{90}{other flat} & \rotatebox{90}{sidewalk} & \rotatebox{90}{terrain} & \rotatebox{90}{manmade} & \rotatebox{90}{vegetation} \\
\midrule
RangeNet++ \cite{milioto2019rangenet++}      & LiDAR          & 65.50 & 66.00   & 21.30   & 77.20 & 80.90 & 30.20                 & 66.80      & 69.60      & 52.10         & 54.20   & 72.20 & 94.10              & 66.60       & 63.50    & 70.10   & 83.10   & 79.80      \\
Cylinder3D \cite{zhu2021cylindrical}    & LiDAR          & 76.10  & 76.40   & 40.30   & 91.20 & 93.80 & 51.30             &78.00      &78.90    & 64.90        &62.10   & 84.40 & 96.80             & 71.60      & 76.40   &75.40   & 90.50   & 87.40      \\
\midrule
TPVFormer$^{ \ast }$~\cite{huang2023tri} & Camera            & 58.45 & 65.99   & 24.50   & 80.88 & 74.28 & 47.04                 & 47.09      & 33.42      & 14.52         & 53.96   & 70.79 & 88.55              & 61.63       & 59.46    & 63.15   & 75.76   & 74.17      \\
\rowcolor[RGB]{220,220,220}{OccNet (ours)}    & Camera            & 60.46 & 66.95   & 32.58   & 77.37 & 73.88 & 37.62                 & 50.87      & 51.45      & 33.69         & 52.20   & 67.08 & 88.72              & 57.99       & 58.04    & 63.06   & 78.91   & 76.97     \\
\bottomrule
\end{tabular}%
}
\vspace{0.5mm}
\caption{\textbf{The performance of OccNet (ResNet101) on nuScenes validation set for LiDAR segmentation task.} OccNet with camera input is comparable with LiDAR based method.Methods with $^{ \ast }$ stands for training form stratch on OpenOcc dataset.}
\label{tab:lidarseg}
\end{table*}

\noindent\textbf{Semantic Scene Completion.}
We compare OccNet with previous state-of-the-art methods for semantic scene completion task in Table~\ref{tab:occ_method} and Figure~\ref{fig:compare_occ_method}.
We reproduce BEVDet4D \cite{huang2022bevdet4d}, BEVDepth \cite{li2022bevdepth} and BEVDet \cite{huang2021bevdet} by replacing the detection head with the scene completion head built on their BEV feature maps, and OccNet outperforms these methods by a large margin as shown in Table~\ref{tab:occ_method}. 
Compared with BEV feature map, our occupancy descriptor is better for the voxel-wise prediction task.
We also compare OccNet with TPVFormer~\cite{huang2023tri}, which is developed for surrounding 3D semantic occupancy prediction task, and our model surpasses it over 3.31 points in terms of mIOU (26.98 vs. 23.67), indicating that occupancy descriptor is better than TPV features for scene representation.
Note that TPVFormer surpasses the OccNet in car, truck and trailer, because samples of these three objects are relatively large in the benchmark and TPVFormer learns better feature on these classes from their sampling strategy.
However, for the objects with small size such as pedestrian and traffic cone, our method can outperform TPVFormer~\cite{huang2023tri} with a large margin of 10 points in Table~\ref{tab:occ_method}.

\noindent\textbf{Occupancy for LiDAR Segmentation.}
Occupancy is a voxelized representation of points in 3D space, and semantic scene completion is equivalent to semantic LiDAR prediction task when $\Delta s \to 0$.
We transfer semantic occupancy prediction to LiDAR segmentation by assigning the point label based on associated voxel label, and then evaluate the model on the mIoU metric.
As reported in Table~\ref{tab:lidarseg}, given camera as input without LiDAR supervision, OccNet can be comparative with the LiDAR segmentation model RangeNet++~\cite{milioto2019rangenet++} in terms of mIoU (60.46 vs. 65.50), and OccNet can even outperforms  RangeNet++ in the IoU of bicycle (32.58 vs. 21.30).
Compared with TPVFormer~\cite{huang2023tri}, OccNet also outperforms it with 2 points in mIoU.

\noindent\textbf{Occupancy for 3D Detection.}
In the scene completion task, the location of foreground object can be coarse regressed, which can help the 3D detection task with 3D box regression.
As shown in Table~\ref{tab:joint_train}, the joint training of scene completion and 3D detection task can improve the detector performance for all our three models, including BEVNet, VoxNet and OccNet, in terms of mAP and NDS.
Note that the voxelized representation of occupancy with $\Delta s=0.5\text{m}$ is too coarse when calculating the metric dependent on the precise center distance and IoU of 3D box, and thus mATE and mASE is a little increased with joint training.

\begin{figure}
  \centering
  \includegraphics[width=\linewidth]{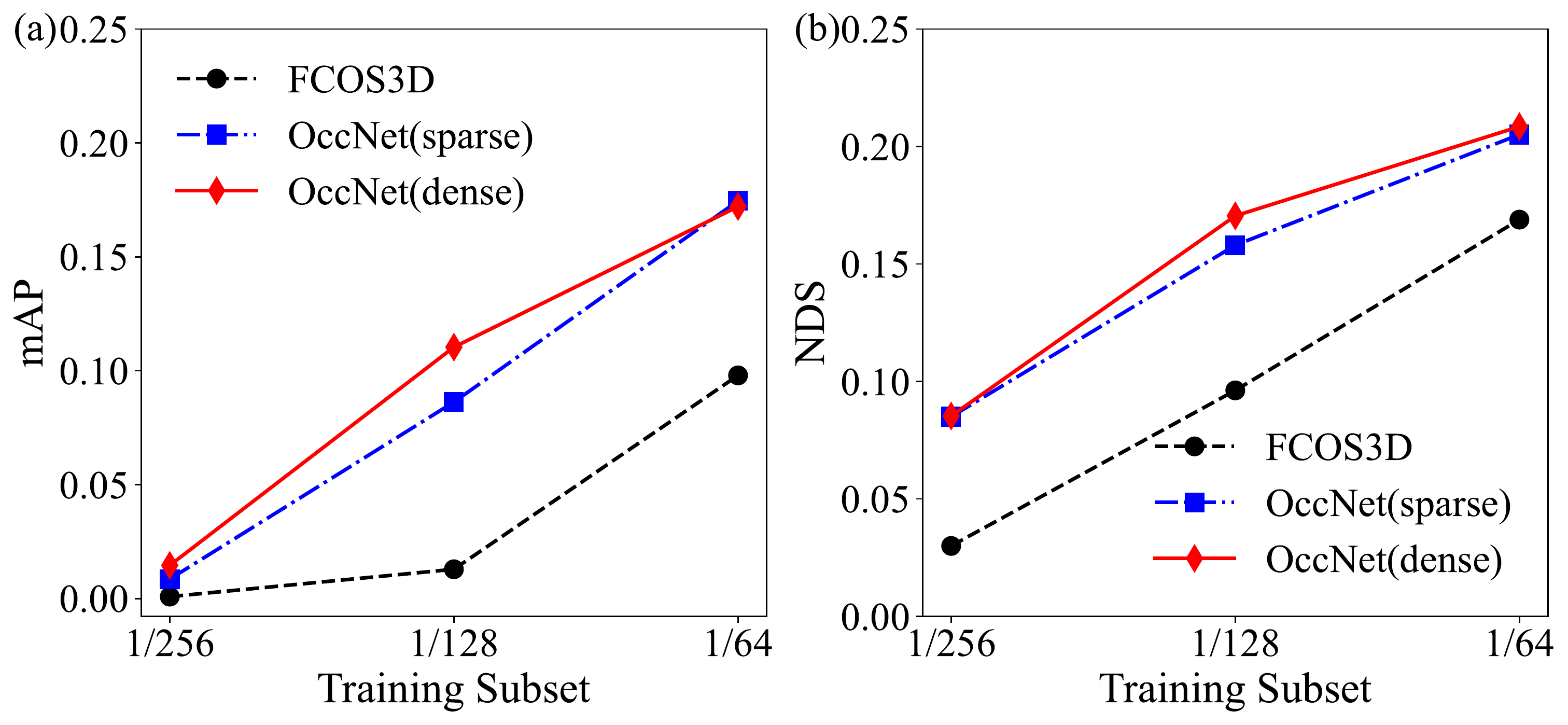}
  \caption{\textbf{The comparison of detector performance using different pretained models and different scale of training dataset.}
  OccNet (sparse) and OccNet (dense) means the OccNet trained on sparse and dense occupancy data respectively. Best view in color.
  }
  \label{fig:pretrain_for_detection}
\end{figure}

\noindent\textbf{Pretrained Occupancy for 3D Detection and BEV segmentation.} The OccNet trained on semantic scene completion task can obtain general representation for 3D space owing to the scene reconstructed in the occupancy descriptor.
Thus, the learned occupancy descriptor can be directly transferred to the downstream 3D perception tasks with model fine-tuning. 
As described in Figure~\ref{fig:pretrain_for_detection}, the model performance on 3D detection with pretained OccNet is superior to that pretrained on FCOS3D~\cite{wang2021fcos3d} detector in different scales of training dataset with the performance gain about 10 points for mAP and NDS.
We also compare the occupancy pretraining and detection pretraining for the BEV segmentation task, indicating that the occupancy pretraining can help BEV segmentation achieve higher IoU in the fine-tuning stage on both semantic and map segmentation as shown in Table~\ref{tab:bev_seg}.


\begin{table}
\setlength{\aboverulesep}{0pt} 
\setlength{\belowrulesep}{0pt} 
\setlength{\tabcolsep}{2pt}
\renewcommand{\arraystretch}{1.2}
\centering
\resizebox{\linewidth}{!}{
\begin{tabular}{l|c|>{\colortable}c|>{\colortable}c|ccccc}
\toprule
Method                    & Joint        & mAP$\uparrow$    & NDS$\uparrow$    & mAOE$\downarrow$   & mAVE$\downarrow$    & mAAE$\downarrow$    & mATE$\downarrow$    & mASE$\downarrow$  \\
\midrule  
\multirow{2}{*}{BEVNet}  &      -         & 0.259 & 0.377  & 0.600 & 0.592  & 0.216 & \textbf{0.828} & \textbf{0.290} \\
                           &     \checkmark          & \textbf{0.271} & \textbf{0.390}   & \textbf{0.578} & \textbf{0.541} & \textbf{0.211} & 0.835 & 0.293 \\
\midrule                          
\multirow{2}{*}{VoxNet} &      -         & 0.271 & 0.380 & 0.603 & 0.616 & 0.219 & 0.832 & \textbf{0.284}   \\
                          &     \checkmark          & \textbf{0.277} & \textbf{0.387} & \textbf{0.586} & \textbf{0.614} & \textbf{0.203} & \textbf{0.828} & 0.285 \\
\midrule
\multirow{2}{*}{OccNet}    &     -         & 0.276 & 0.382 & 0.655 & 0.588 & 0.209 & \textbf{0.817} & \textbf{0.290}\\
                           &      \checkmark             & \textbf{0.276}  & \textbf{0.390} & \textbf{0.585} & \textbf{0.570} & \textbf{0.190} & 0.842 & 0.292 \\
\bottomrule
\end{tabular}
}
\vspace{0.5mm}
\caption{\textbf{Joint training of 3D occupancy and 3D detection.} Results reported on nuScenes validation set show that joint training of 3D occupancy and 3D detection can help the latter task.}
\label{tab:joint_train}
\end{table}

\begin{table}[t]
  \small
  \centering    
  \resizebox{\linewidth}{!}{
  \begin{tabular}{l|>{\colortable}c|cccc}
    \toprule
    Task & Main value & Drivable area & Lane & Vehicle & Pedestrian \\
    \midrule
    Det &  18.18 & 44.59 & 13.62 & 12.29 & 2.21 \\
    Occ &  \textbf{19.17} & \textbf{47.21} & \textbf{13.83} & \textbf{12.91} & \textbf{2.74}\\
    \bottomrule
  \end{tabular}
  }
  \vspace{0.5mm}
  \caption{\textbf{Different pretraining tasks for BEV segmentation.} Occupancy task can help BEV segmentation task achieve higher IoU.}
  \vspace{-0.5mm}
  \label{tab:bev_seg}
\end{table}

\begin{figure}[t]
  \centering
    \includegraphics[width=0.95 \linewidth]{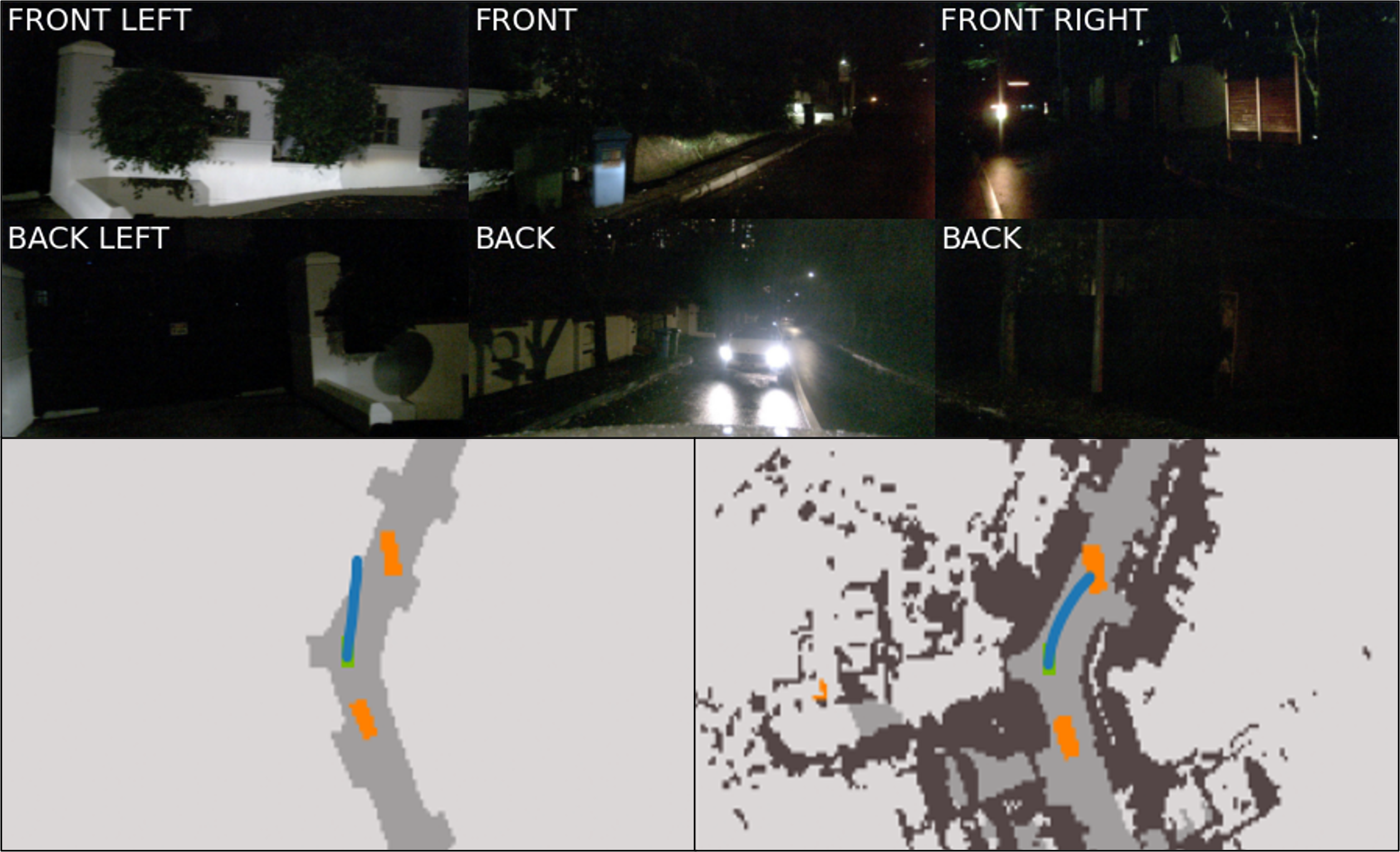}
  \caption{\textbf{Visualization of planning.} The blue line represents the planned trajectory, and the lower figures are rasterisation results of bounding box and occupancy, respectively.}
  \label{fig:planning}
\end{figure}

\noindent\textbf{Occupancy for Planning.}
With the prediction results from upstream tasks, i.e., bounding box and occupancy, the final trajectory can be obtained through a cost filter and a GRU refinement module~\cite{st-p32022} with the BEV segmentation inputs. To obtain these segmentation results, we rasterize the outputs of our OccNet in BEV space.
We compare the rasterisation results of bounding box and occupancy by using the predictions from OccNet. We also compare our results with the direct segmentation from ST-P3~\cite{st-p32022}. For a fair comparison, we follow the same setup as ST-P3 with only vehicle and pedestrian classes kept. We also add the ground truth rasterisation inputs for better comparison. As shown in Table~\ref{tab:planning}, the best performance can be obtained by using ground truth of occupancy to filter trajectories. For predicted results, the collision rate can be reduced by 15\% - 58\% based on the occupancy prediction from OccNet.
We also conduct the experiment using all 16 classes of occupancy, which shows that full classes of occupancy can bring the performance improvement on L2 distance. 
As shown in the Figure~\ref{fig:planning}, planning with full classes of occupancy can make decisions within the feasible areas to avoid collisions from the background objects.

\begin{table}[t]
  \small
  \centering    
  \resizebox{\linewidth}{!}{
  \begin{tabular}{l|ccc|ccc}
    \toprule
    \multirow{2}*{Input} &  \multicolumn{3}{c|}{Collision ($\%$)$\downarrow$} & \multicolumn{3}{c}{L2 (m)$\downarrow$}\\
    & 1s & 2s & 3s & 1s & 2s & 3s \\
    \midrule
    Bbox GT & 0.23 & 0.66 & 1.50 & 1.32 & 2.16 & 3.00 \\
    Occupancy GT & \textbf{0.20} & \textbf{0.56} & \textbf{1.30} & \textbf{1.29} & \textbf{2.13} & \textbf{2.98} \\ 
    \midrule
    Segmentation pred. \cite{st-p32022} & 0.50 & 0.88 & 1.49 &1.39 & 2.21 & 3.02 \\
    Bbox pred. (OccNet) & 0.27 & 0.68 & 1.59 &1.32 & 2.17 & 3.03 \\
    \rowcolor[RGB]{220,220,220}Occupancy pred. (OccNet) & \textbf{0.21} & \textbf{0.55} & \textbf{1.35} & 1.31 & 2.18 & 3.07 \\
    \rowcolor[RGB]{220,220,220}Occupancy pred. (OccNet, full) & \textbf{0.21} & 0.59 & 1.37 &\textbf{1.29} & \textbf{2.13} & \textbf{2.99} \\
    \bottomrule
  \end{tabular}
  }
  \vspace{0.5mm}
  \caption{\textbf{Planning results with different scene representations.} Occupancy representation helps the planning task achieve a lower collision rate and more accurate L2 distance in all time intervals.}
  \label{tab:planning}
\end{table}

\subsection{Discussion}

\noindent\textbf{Model Efficiency.} In Table~\ref{tab:efficiency}, we compare the performance of different models in the semantic scene completion task. Compared with BEVNet and VoxelNet, OccNet can obtain the best performance in terms of mIOU and $\text{IoU}_{geo}$ with efficiency and effectiveness.

\begin{table}[t]
\small 
\centering
\resizebox{0.8\linewidth}{!}{
\begin{tabular}{l|cc|ccc}
\toprule
Method   & mIOU & $\text{IoU}_{geo}$ & Params$\downarrow$   & Memory$\downarrow$ & FPS$\uparrow$ \\
\midrule
BEVNet   & 17.37  & 36.11& 39M & 8G   &   4.5  \\
VoxelNet & 19.06  & 37.59 & 72M & 23G  &  1.9   \\
\rowcolor[RGB]{220,220,220}OccNet   & \textbf{19.48}  & \textbf{37.69} & 40M & 18G  &  2.6  \\
\bottomrule
\end{tabular}%
}
\vspace{0.5mm}
\caption{\textbf{Efficiency and performance analysis with model structure.} The evaluation is measured on a V100 GPU.}
\label{tab:efficiency}
\end{table}

\begin{figure}
  \centering
  \includegraphics[width=\linewidth]{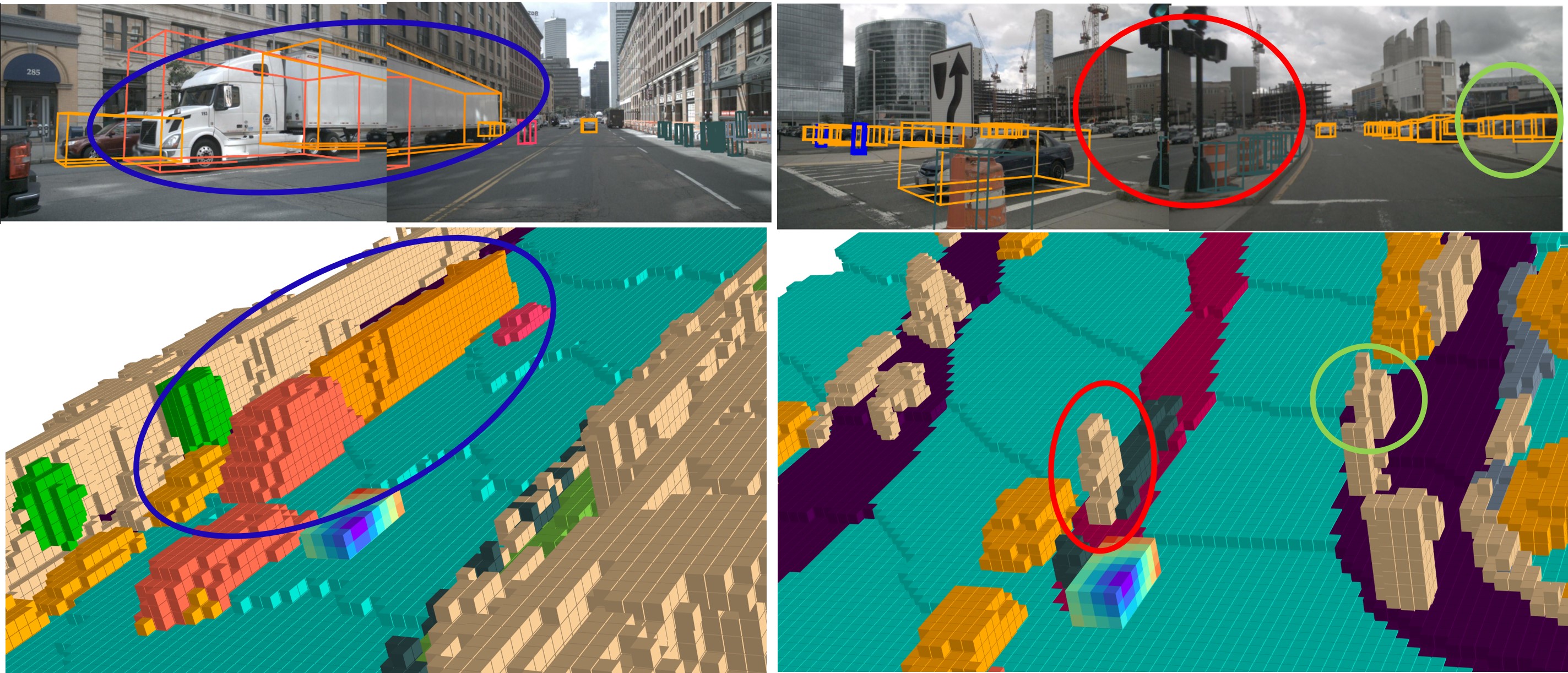}
  \caption{\textbf{Visualization of 3D box and occupancy prediction.}}
  \label{fig:irregular}
\end{figure}

\noindent\textbf{Irregular Object.} Representing the irregular object such as construction vehicle with 3D box or the background stuff such as traffic sign is difficult and inaccurate as indicated in Figure~\ref{fig:irregular}. We transform 3D box into voxel to compare the 3D detection and occupancy task on irregular object in Table~\ref{tab:irregular}, verifying that occupancy can describe the irregular object better. 
To study the effect of voxel size, we also generate the dataset with $\Delta s=0.25\text{m}$. With the decrease of $\Delta s$ from 0.5m to 0.25m, the performance gap between 3D box and occupancy increases because the finer granularity can better depict the irregular object.

\begin{table}
\centering
\resizebox{0.8\linewidth}{!}{%
\begin{tabular}{l|c|c|ccc}
\toprule
Task &  $\Delta s$  & $\text{mIoU}_{ir}$ & truck & trailer & cons. veh. \\
\midrule
Det  & 0.5m  & 15.92                       & 23.19 & 9.57    & 15.00      \\
Occ  & 0.5m  & \textbf{18.13 (\color{red}{+2.21}})                      & \textbf{25.59} & \textbf{12.29}   & \textbf{16.51} \\
\midrule
Det  & 0.25m & 9.90 & 14.84 & 5.10 & 9.75 \\
Occ  & 0.25m & \textbf{13.41 (\color{red}{+3.51})} & \textbf{18.85} & \textbf{7.14} & \textbf{14.25} \\
\bottomrule
\end{tabular}%
}
\vspace{0.5mm}
\caption{\textbf{The comparison of detection task and occupancy task on the recognition of irregular object}. $\text{mIoU}_{ir}$ denotes mean IoU of truck, trailer and construction vehicle.}
\label{tab:irregular}
\end{table}

\noindent\textbf{Dense v.s. Sparse Occupancy.}
Compared with sparse occupancy, dense occupancy can help depict the complete geometry of background and foreground object in detail as shown in Figure~\ref{fig:compare_gt}.
Intuitively, dense occupancy is better for 3D perception and motion planning owing to more abundant information input.
We validate that model pretrained on dense occupancy can benefit the downstream 3D detection task more as shown in Figure~\ref{fig:pretrain_for_detection}.

\section{Conclusion}

We dive into the potential of the 3D occupancy as scene representation and propose a general framework OccNet to evaluate the idea. The experiments on various downstream tasks validate the effectiveness of our method. The OpenOcc benchmark with dense and high-quality labels is also provided for community. 

\noindent\textbf{Limitations and future work}. Currently, the annotation is still based on the well-established dataset. Utilizing self-supervised learning to further reduce the human-annotation cost is a valuable direction. We hope occupancy framework can be the foundation model of autonomous driving.



{\small
\bibliographystyle{ieee_fullname}
\bibliography{egbib}
}

\clearpage 

\input{sup_arxiv}

\end{document}

%% file: sections/abstract_smch.tex
Human driver can easily describe the complex traffic scene by visual system.
Such an ability of precise perception is essential for driver's planning.
%
To achieve this,
a geometry-aware representation that quantizes the physical 3D scene into structured grid map with semantic labels per cell, termed as \emph{3D Occupancy}, would be desirable.
Compared to the form of bounding box, a key insight behind occupancy is that it could capture the fine-grained details of critical obstacles in the scene, and thereby facilitate subsequent tasks.
Prior or concurrent literature mainly concentrate on a single scene completion task, 
where we might argue that the potential of this occupancy representation might obsess broader impact.
%
In this paper, we propose OccNet, a multi-view vision-centric pipeline with a cascade and temporal voxel decoder 
to reconstruct 3D occupancy.
At the core of OccNet is a general occupancy embedding to represent 3D physical world. Such a descriptor could be applied towards a wide span of driving tasks, including detection, segmentation and planning.
To validate the effectiveness of this new representation and our proposed algorithm,
we propose OpenOcc, the first dense high-quality 3D occupancy benchmark built on top of nuScenes.
%
%
%
%
Empirical experiments show that there are evident performance gain across multiple tasks,
e.g.,
%
motion planning could witness a collision rate reduction 
by 15\%-58\%,
demonstrating the superiority of our method.

%% file: sup_arxiv.tex
\appendix
\noindent\textbf{\large{Appendix}}

We put the detail of evaluation metrics in the supplementary materials, along with more related work, visualization, implementation / training detail and more ablation of OccNet, and detail about BEVNet, VoxelNet and OpenOcc post-processing.

\section{Evaluation Metrics}

\noindent\textbf{Semantic Scene Completion (SSC) Metric.} For the scene completion task, we predict the semantic label of each voxel in 3D space. The evaluation metric is defined by mean intersection-over-union (mIoU) over all classes:
\begin{equation}
    \text{mIoU} = \frac{1}{C}\sum_{c=1}^{C} \frac{\text{TP}_c}{\text{TP}_c+\text{FP}_c+\text{FN}_c},
\label{eq:miou}
\end{equation}
where $C=16$ is the class num in the benchmark, $\text{TP}_c$, $\text{FP}_c$ and $\text{FN}_c$ represent true positive, false positive, and false negative predictions for class $c$, respectively. 
In addition, we consider the class-agnostic metric $\text{IoU}_{geo}$ to evaluate the geometrical reconstruction quality of scene.

\noindent\textbf{3D Object Detection Metric.} We use the official evaluation metrics for the nuScenes datasets~\cite{caesar2020nuscenes}, including nuScenes detection score (NDS), mean average precision (mAP),  average
translation error (ATE), average scale error (ASE), average orientation error (AOE), average velocity error (AVE)
and average attribute error (AAE).

\noindent\textbf{Motion Planning Metric.}
For planning evaluation, we follow the metrics in ST-P3~\cite{st-p32022}. In detail, L2 distance is calculated by the planning trajectory and the ground-truth trajectory for the regression accuracy, and collision rate (CR) to other vehicles and pedestrians is applied for the safety of future actions.

\section{More Related Work}

\noindent\textbf{BEV segmentation} 
%
\cite{zhou2022cross,li2021hdmapnet} implicitly squeezes the height information into each cell in BEV map. However, in some challenging urban settings, explicit height information is necessary to capture entities above the ground, \textit{e.g.} traffic lights and overpass.
%
As an alternative,
3D occupancy is 3D geometry-aware.

\section{Implementation Detail of OccNet}

\paragraph{Backbone and Multi-scale Features.}
Following previous works \cite{li2022bevformer, wang2021fcos3d}, We adopt ResNet101 \cite{he2016resnet} as the backbone with FPN \cite{lin2017feature} to extract the multi-scale features from multi-view images.
We use the output features from stages $S_3$, $S_4$, and $S_5$ from ResNet101, where $S_n$ means the downsampling factor is $1/2^n$ with the feature dimension $C_n=256\times 2^{n-2}$.
In the FPN, the features are aggregated and transforms to three levels 
with sizes of 1/16, 1/32, 1/64 and the dimension of $C_n=256$.

\paragraph{BEV Encoder.}
The BEV encoder follows the structure of BEVFormer \cite{li2022bevformer},
where the multi-scale features from FPN are transformed into the BEV feature.
The BEV encoder includes 2 encoder layers with the temporal self-attention and spatial cross-attention. Then the BEV query $Q_t$ gradually refines in the encoder layers with spatial-temporal-transformer mechanism to learn the scene representation in BEV space. 

\paragraph{Feature Transformation in Voxel Decoder.}
To lift the voxel feature $V_{t, i}^{'} \in \mathbb{R}^{Z_i \times H \times W \times C_i}$ to $V_{t, i+1}^{'} \in \mathbb{R}^{Z_{i+1} \times H \times W \times C_{i+1}}$, we use the MLP to transfer the feature dimension from $Z_i \times C_i$ to $Z_{i+1} \times C_{i+1}$.
To implement the spatial cross attention for $V_{t, i}^{'}$, the multi-scale image features from FPN with dimension of $C_n=256$ are transformed into dimension of $C_i$ utilizing the MLP.

\paragraph{Training Strategy.}

Following previous works \cite{li2022bevformer, wang2021fcos3d}, we train OccNet 24 epochs with a learning rate of $2 \times 10 ^{-4}$, a batchsize of 1 per GPU with six images, and AdamW optimizer \cite{loshchilov2017decoupled} with a weight decay of 
$1 \times 10^{-2}$.
For the implementation of downstream tasks, all the perception
tasks (except BEV segmentation) are trained at once, and
the others are fine-tuned based on the frozen tasks.

\paragraph{Details of VoxelNet and BEVNet.}
Different from the OccNet with cascaded feature map, we construct the VoxelNet and BEVNet with single-scale feature map.
In detail, VoxelNet uses voxel queries $Q_{voxel} \in  \mathbb{R}^{4 \times H \times W}$ to construct the voxel feature map
$F_{voxel} \in \mathbb{R}^{4 \times H \times W \times C_1}$
from the image feature using 3D-DA directly, and expands it to full-scale occupancy $V \in \mathbb{R}^{16 \times H \times W \times C_2}$ using fully connected layer.
%
BEVNet generates BEV feature $F_{bev} \in \mathbb{R}^{ H \times W \times C_1}$ as in BEVFormer and reshapes it to voxel feature $V \in \mathbb{R}^{16 \times H \times W \times C_2}$ directly.
%
Here $C_1$ and $C_2$ stand for the number of channels.
Both VoxelNet and BEVNet adopt temporal context fusion accordingly.

\section{More Detail about OpenOcc}

\begin{figure}
  \centering
    \includegraphics[width=0.95 \linewidth]{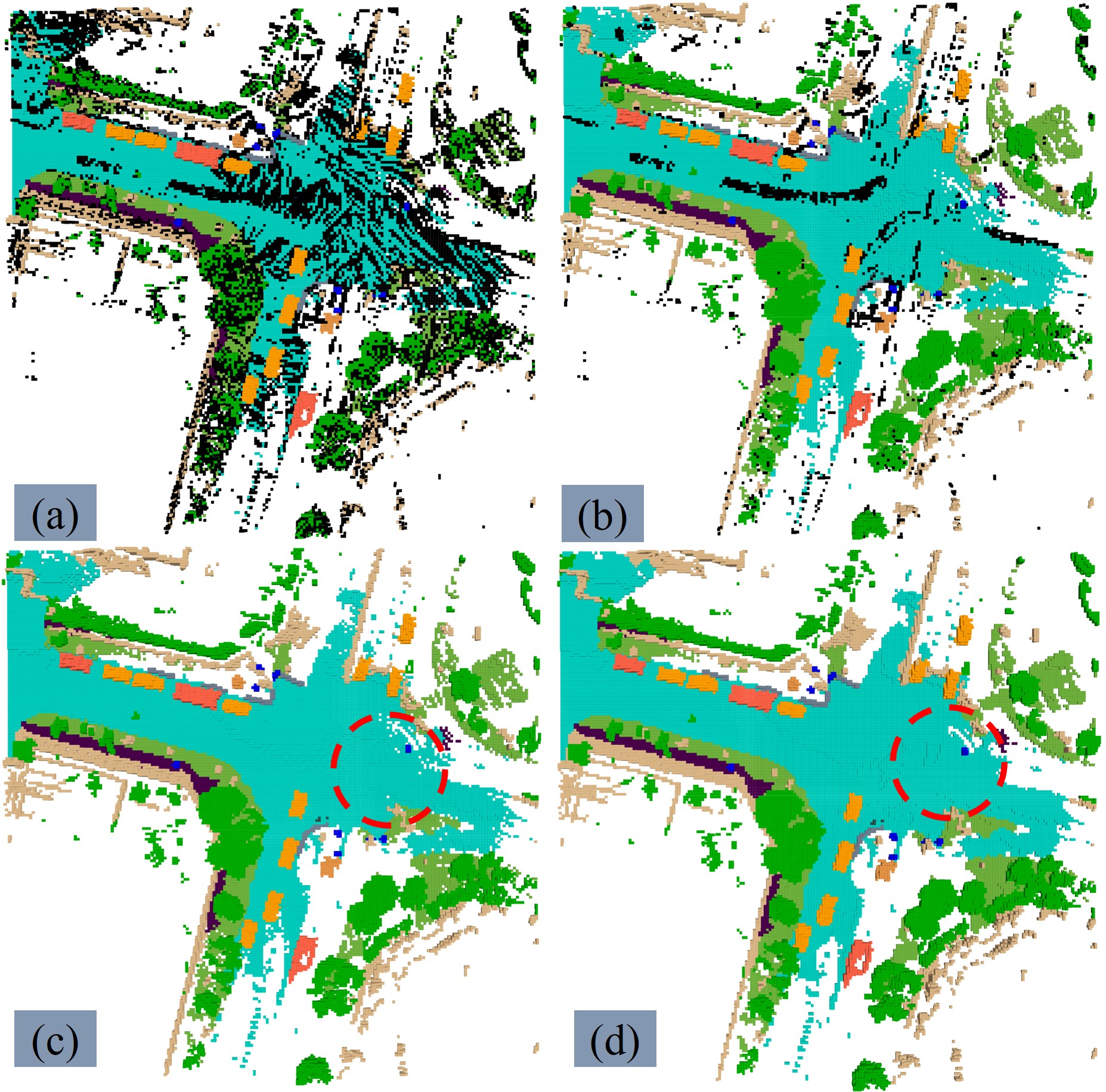}
  \caption{\textbf{The generation process of our occupancy data.} (a) Generating the occupancy data based on objects points and partial background points with label, where the black points denotes the unknown background points from the intermediate frame.
  (b) Annotating partial unknown background points based on generated occupancy data.
  (c) Removing the remaining unknown background points which are regarded as noise.
  (d) Postprocessing the occupancy data to ensure the completeness of the scene, such as fill the hole, denoted by the red dashed box.
  }
  \label{fig:occ_gt_pipeline}
\end{figure}

\paragraph{Accumulation of Foreground objects.}
To accumulate the foreground object, we split the LiDAR points into object points and background points. However, the 3D box annotation of intermediate frame is not provided in the nuScenes dataset \cite{caesar2020nuscenes}. We approximately annotate the 3D box using the linear interpolation based on two adjacent key frames, then we can accumulate dense object points with available intermediate LiDAR points.

\paragraph{Dataset Generation Pipeline.}
With accumulated dense background points and foreground object points, we
generate the occupancy data following the pipeline as shown in Figure~\ref{fig:occ_gt_pipeline}. We gradually fine tune the occupancy data and obtain the 3D occupancy benchmark with dense and
high-quality annotations in Figure~\ref{fig:occ_gt_pipeline}(d).

\begin{figure}[t]
    \centering
    \includegraphics[width=.95\linewidth]{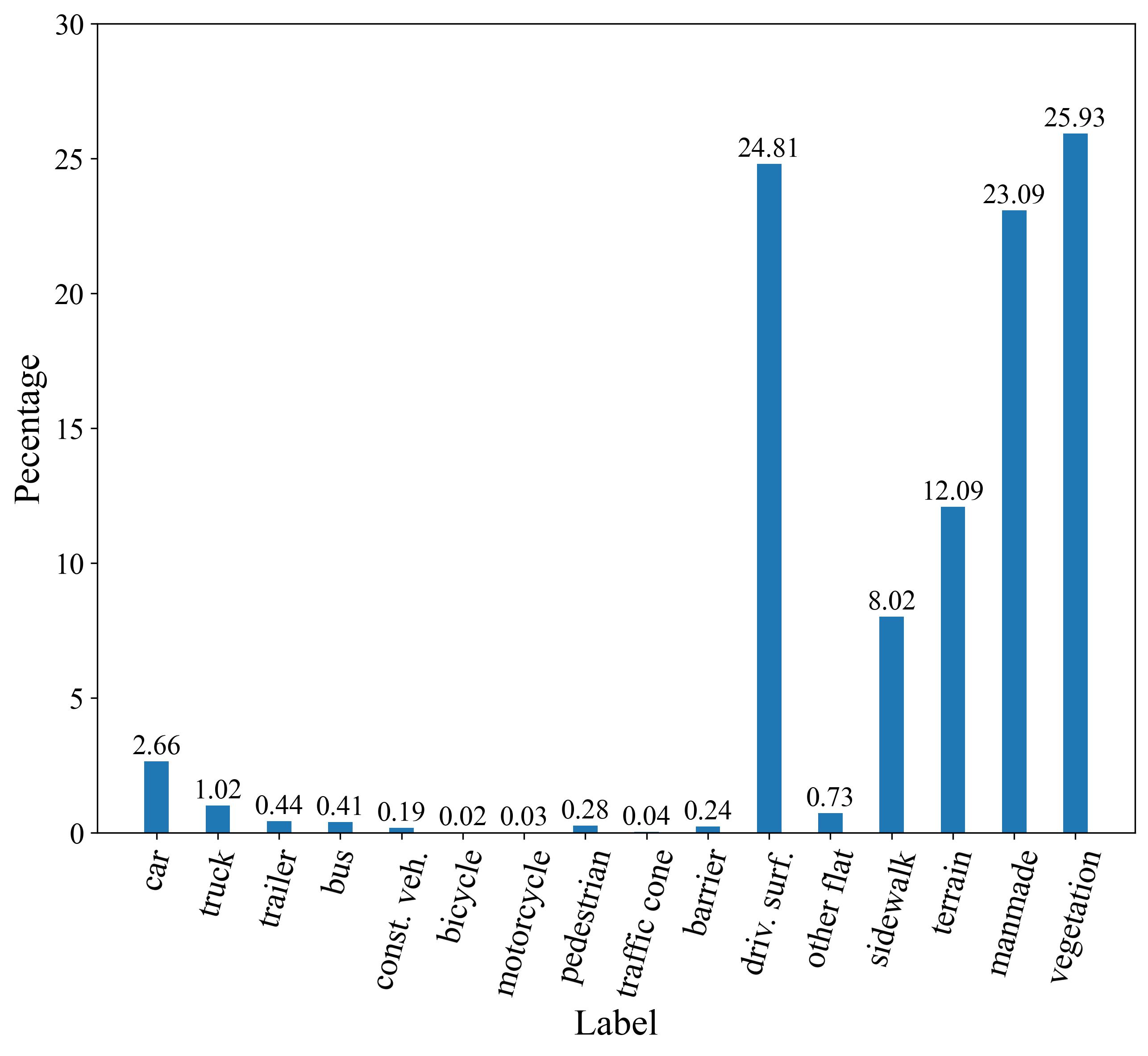}
    \caption{\textbf{The distribution of occupancy classes in the OpenOcc benchmark.} We notice that the background stuff is the majority in 3D occupancy data.}
    \label{fig:statistics_occ}
\end{figure}

\paragraph{Dataset Statistics.}
We annotate 16 classes in 34149 frames for all 700 training and 150 validation scenes with over 1.4 billion voxels.
The label distribution of 16 classes is shown in Figure~\ref{fig:statistics_occ}, indicating great diversity in the benchmark.
There exists a significant class imbalance phenomenon in the dataset, for example, where the 10 foreground objects only account for 5.33\% of the total labels, especially the bicycle and motorcycle, which account for 0.02\% and 0.03\%, respectively.

\begin{figure}[t]
    \centering
    \includegraphics[width=.95\linewidth]{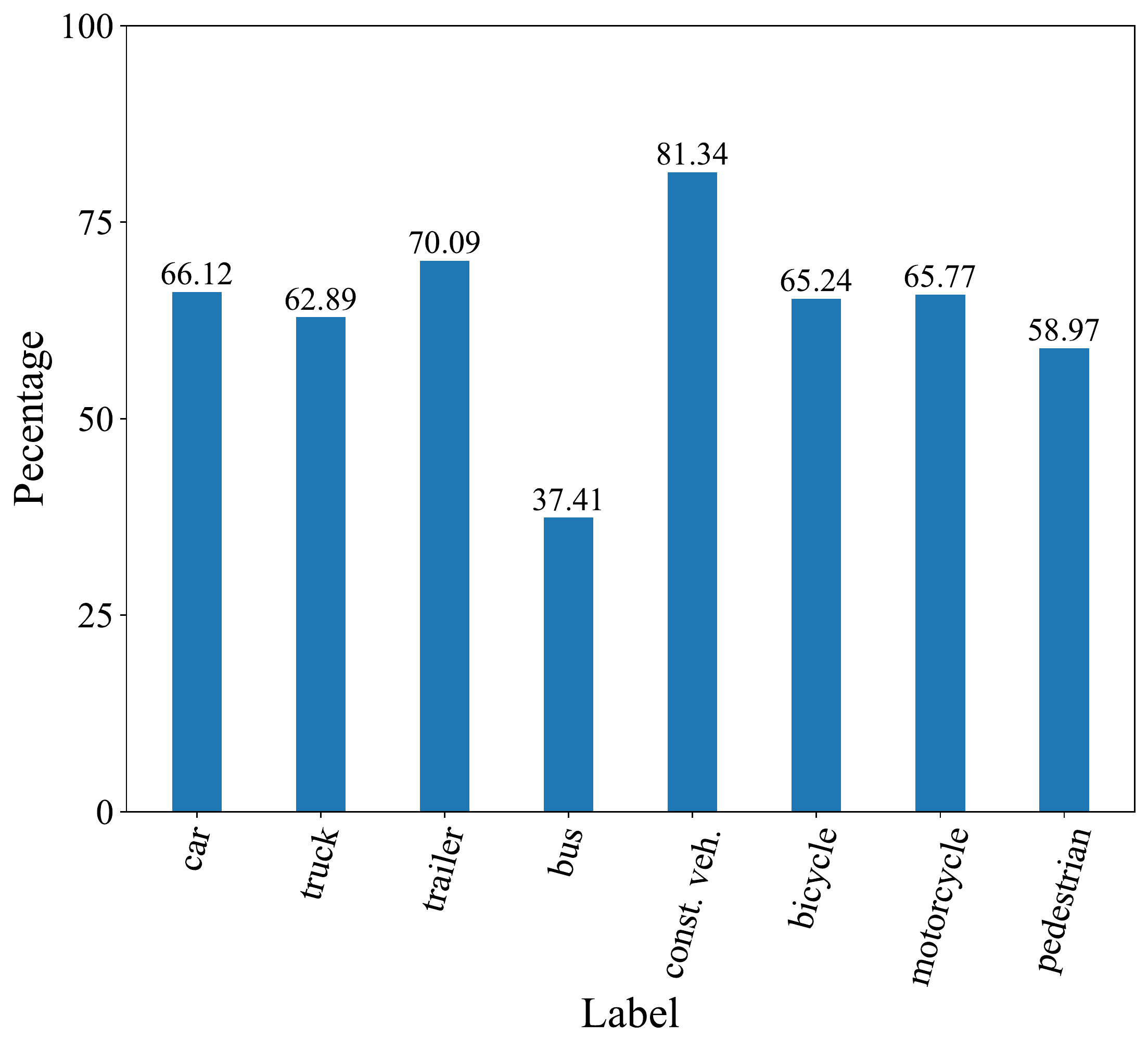}
    \caption{\textbf{The percentage of occupancy with velocity for each foreground object.} For 10 foreground objects in the benchmark, we only consider the 8 movable classes.}
    \label{fig:statistics_flow}
\end{figure}

We provide the additional flow annotation of eight foreground objects, which is helpful for the downstream task such as motion planning.
We split the object into moving state and stationary state based on the velocity threshold $v_{th}=\text{0.2m/s}$, and the percentage of moving object for each class is given in Figure~\ref{fig:statistics_flow}.
Note that the percentage of moving foreground object is over 50\%, indicating the significance of motion information in the autonomous driving scenes.

\section{More Experiments}
\paragraph{Ablations on Frame Number for Temporal Self-Attention.}
We investigate the effect of frame number applied for temporal self-attention during training.
From Table~\ref{tab:frame_number_ssc} and Table~\ref{tab:frame_number_lidarseg},
we find that increasing temporal frames results in better performance, which slows down
until a threshold of four frames is reached. Meanwhile, 
insufficient previous frames would hurt the performance to
some extent.

\begin{table}[h!]
\centering
\scalebox{0.8}{
\begin{tabular}{l|ccccc}
\toprule
  \#Num & \text{IoU}$_{geo}$$\uparrow$ & mIoU$\uparrow$ & barrier$\uparrow$ & bicycle$\uparrow$ & bus$\uparrow$ \\ 
  \midrule
  0 & 37.49 & 19.21 & 20.07 & 4.70 & 24.11 \\
  1 & 36.89 & 18.35 & 18.77 & 4.51 & 21.66\\
  2$^{ \ast }$ & {37.69} & {19.48} & {20.63} & 5.52 & {24.16} \\ 
  3 & 38.36 & 20.30 &  21.39 & 6.47 & 24.65\\
  4 & 39.21 & \textbf{20.81} &  \textbf{22.30} & 5.66 & \textbf{25.13}\\ 
  9 & \textbf{39.36} & 20.68 &  20.75 & \textbf{7.83} & 24.79\\ 
 \bottomrule
\end{tabular}}
\vspace{0.5mm}
  \caption{\textbf{The effect of historical frames on the semantic scene completion task using OccNet with ResNet50 backbone.} The ``\#Num" denotes the historical frame number used during training. $^{ \ast }$ stands for number used in the main paper.}
\label{tab:frame_number_ssc}
\end{table}

\begin{table}[h!]
\centering
\scalebox{0.8}{
\begin{tabular}{l|cccccc}
\toprule
  \#Num & mIoU$\uparrow$ & barrier$\uparrow$ & bicycle$\uparrow$ & bus$\uparrow$ & car$\uparrow$ & truck$\uparrow$ \\ 
  \midrule
  0 & 53.82 & 59.02 & 24.05 & 67.61 & 69.59 &  59.38 \\
  1 & 48.41 & 57.45 & 21.52 & 57.71 & 67.26 & 44.60\\
  2  & 52.33 & 61.41 & 26.07 & 73.97 &70.56 & 52.64\\ 
  3 & 53.49 & 60.98 & 24.45 &70.20 & 69.37 & 56.84 \\
  4 & \textbf{54.59} & \textbf{62.09} & 21.06 & 75.05 & 70.20 & 59.40 \\
  9 & 54.35 & 60.04 & \textbf{30.83} & \textbf{75.49} & \textbf{71.02} & \textbf{61.63}\\
 \bottomrule
\end{tabular}}
\vspace{0.5mm}
  \caption{\textbf{The effect of historical frames on the LiDAR segmentation task using OccNet with ResNet50 backbone.}
  The ``\#Num" denotes the historical frame number used during training.}
  \label{tab:frame_number_lidarseg}
\end{table}

\paragraph{Evaluation on Occupancy Metrics for Planning.} We utilize the ground truth of occupancy as the metrics to evaluate the planning model, instead of the bounding box of vehicles and pedestrians. Specifically, all of foreground occupancy voxels and four classes of background occupancy voxels, i.e., other flat, terrain, manmade, and vegetation, are calculated collision rate with trajectory. As shown in Table~\ref{tab:occupancy metrics}, using occupancy as input for planning model is still more advantageous in most of the intervals under this metrics. In the future research, a specific design of cost function for occupancy input may further improve the performance of planning.

 \begin{table}[h!]
  \small
  \centering    
  \resizebox{\linewidth}{!}{
  \begin{tabular}{l|ccc|ccc}
    \toprule
    \multirow{2}*{Input} &  \multicolumn{3}{c|}{Collision ($\%$)$\downarrow$} & \multicolumn{3}{c}{L2 (m)$\downarrow$}\\
    & 1s & 2s & 3s & 1s & 2s & 3s \\
    \midrule
    Bbox GT & 1.66 & 2.88 & 4.37 & 1.33 & 2.18 & 3.03 \\
    Occupancy GT & \textbf{1.63} & \textbf{2.85} & \textbf{4.29} & \textbf{1.29} & \textbf{2.13} & \textbf{2.99} \\ 
    \midrule
    Bbox pred. (OccNet) & 1.75 & \textbf{2.85} & 4.37 &1.33 & 2.17 & 3.04 \\
    Occupancy pred. (OccNet) & \textbf{1.68} & 2.94 & \textbf{4.32} & \textbf{1.30} & \textbf{2.15} & \textbf{3.02} \\
    \bottomrule
  \end{tabular}
  }
  \vspace{0.5mm}
  \caption{\textbf{Planning results with different scene representations under occupancy metrics.} Occupancy representation is still more advantageous most of the intervals.}
  \label{tab:occupancy metrics}
\end{table}

\paragraph{Pre-training for planning.} As evaluating the pre-trained model on 3D detection and BEV segmentation tasks in the main paper, we further compared the impact on the downstream planning task. Specifically, the perception module of ST-P3~\cite{st-p32022} is replaced by pre-trained OccNet, and the planning module is fine-tuned. Unfortunately, the pre-training on OccNet does not provide an advantage for planning as shown in Table~\ref{tab:planning_pretrain}. Therefore, combined with the experiment of planning in the main paper, we should directly apply the scene completion results of occupancy in the planning task instead of these pre-trained features.

\begin{table}[h!]
  \scriptsize
  \centering    
  \resizebox{\linewidth}{!}{
  \begin{tabular}{l|ccc|ccc}
    \toprule
    \multirow{2}*{Input} &  \multicolumn{3}{c|}{Collision ($\%$)$\downarrow$} & \multicolumn{3}{c}{L2 (m)$\downarrow$}\\
    & 1s & 2s & 3s & 1s & 2s & 3s \\
    \midrule
    Det & \textbf{0.38} & \textbf{0.40} & \textbf{0.82} & \textbf{0.85} & \textbf{1.18} & \textbf{1.57} \\
    Occ & 0.47 & 0.68 & 1.03 & 0.93 & 1.26 & 1.70 \\ 
    \bottomrule
  \end{tabular}
}
  \vspace{0.5mm}
  \caption{\textbf{Different pretraining tasks for planning.} Pretrained features from occupancy do not directly bring performance benefits to planning. }
  \label{tab:planning_pretrain}
\end{table}

\paragraph{Ablations in Semantic Scene Completion.}
Table~\ref{tab:occ_ablation} shows the comparison of BEVNet, VoxelNet, OccNet in the task of semantic scene completion.
We can see that the design of cascaded voxel structure can help learn a bettern occupancy descriptor to represent the 3D space.

\begin{table*}[ht!]
\small
\setlength{\tabcolsep}{3.5pt}
\centering
\resizebox{\textwidth}{!}{
\begin{tabular}{@{}l|c|c|>{\colortable}c|cccccccccccccccc@{}}
\toprule
Method                  & Backbone  & $\text{IoU}_{geo}$   & mIoU  &\rotatebox{90}{barrier} & \rotatebox{90}{bicycle} &  \rotatebox{90}{bus}   & \rotatebox{90}{car}   & \rotatebox{90}{const. veh.} & \rotatebox{90}{motorcycle} & \rotatebox{90}{pedestrian} & \rotatebox{90}{traffic cone} & \rotatebox{90}{trailer} & \rotatebox{90}{truck} & \rotatebox{90}{driv. surf.} & \rotatebox{90}{other flat} & \rotatebox{90}{sidewalk} & \rotatebox{90}{terrain} & \rotatebox{90}{manmade} & \rotatebox{90}{vegetation} \\
\midrule
BEVNet          & ResNet50  & 36.11                   & 17.37                    & 14.02                       & 5.07                        & 20.85                   & 24.94                   & 8.64                                      & 7.75                           & 12.8                           & 8.93                              & 10.21                       & 16.02                     & 44.41                                  & 14.42                           & 23.87                        & 27.76                       & 13.73                       & 24.49                          \\
VoxelNet        & ResNet50  & 37.59                   & 19.06                    & 19.31                       & 6.25                        & 22.16                   & 26.89                   & 9.96                                      & 6.91                           & 12.70                           & 6.27                              & 9.43                        & 16.96                     & 46.7                                   & 23.31                           & 26.04                        & 29.08                       & 16.52                       & 26.46                          \\
\rowcolor[RGB]{220,220,220} OccNet      & ResNet50  & 37.69                   & 19.48                    & 20.63                       & 5.52                        & 24.16                   & 27.72                   & 9.79                                      & 7.73                           & 13.38                          & 7.18                              & 10.68                       & 18.00                        & 46.13                                  & 20.60                            & 26.75                        & 29.37                       & 16.90                        & 27.21                          \\
\midrule
BEVNet         & ResNet101 & 40.15                   & 24.62                    & 26.39                       & 15.79                       & 32.07                   & 35.83                   & 11.93                                     & 19.72                          & 19.75                          & 15.38                             & 12.82                       & 23.90                      & 49.16                                  & 21.52                           & 30.57                        & 31.39                       & 18.99                       & 28.71                          \\
VoxelNet        & ResNet101 & 40.73                   & 26.06                    & 27.98                       & 15.95                       & 32.31                   & 36.15                   & 14.88                                     & 20.55                          & 20.72                          & 16.52                             & 15.13                       & 25.94                     & 49.07                                  & 27.82                           & 31.04                        & 32.43                       & 20.45                       & 29.99                          \\
\rowcolor[RGB]{220,220,220} OccNet & ResNet101 & \textbf{41.08}          & \textbf{26.98}           & \textbf{29.77}              & \textbf{16.89}              & \textbf{34.16}          & \textbf{37.35}          & \textbf{15.58}                            & \textbf{21.92}                 & \textbf{21.29}                 & \textbf{16.75}                    & \textbf{16.37}              & \textbf{26.23}            & \textbf{50.74}                         & \textbf{27.93}                  & \textbf{31.98}               & \textbf{33.24}              & \textbf{20.8}               & \textbf{30.68}                       \\
\bottomrule
\end{tabular}%
}
\vspace{0.5mm}
\caption{\textbf{Ablation in semantic scene completion with different models.} OccNet is superior to BEVNet and VoxelNet in performance.
}
\label{tab:occ_ablation}
\end{table*}

\paragraph{Effect of Voxel Resolution on LiDAR Segmentation.}
We voxelize the 3D space with the resolution $\Delta s \in \{1.0 \text{m}, 0.5 \text{m}, 0.25 \text{m}\}$ to investigate the effect of voxel resolution  on LiDAR segmention.
Since we transfer semantic occupancy
prediction to LiDAR segmentation by assigning the
point label based on associated voxel label, the performance of LiDAR segmention will increase with the decrease of $\Delta s$ as shown in Table~\ref{tab:lidarseg}.
OccNet with camera input can achieve the performance of
LiDAR based method with $\Delta s \to 0$.

\begin{table*}[ht!]
\small
\setlength{\tabcolsep}{3.5pt}
\centering
\resizebox{\textwidth}{!}{%
\begin{tabular}{@{}l|c|c|cccccccccccccccc@{}}
\toprule
Method                        & $\Delta s$(m) & mIOU  &\rotatebox{90}{barrier} & \rotatebox{90}{bicycle} &  \rotatebox{90}{bus}   & \rotatebox{90}{car}   & \rotatebox{90}{const. veh.} & \rotatebox{90}{motorcycle} & \rotatebox{90}{pedestrian} & \rotatebox{90}{traffic cone} & \rotatebox{90}{trailer} & \rotatebox{90}{truck} & \rotatebox{90}{driv. surf.} & \rotatebox{90}{other flat} & \rotatebox{90}{sidewalk} & \rotatebox{90}{terrain} & \rotatebox{90}{manmade} & \rotatebox{90}{vegetation} \\
\midrule
OccNet & 1.00  & 46.60          & 52.78            & 21.04            & 65.94        & 62.45        & 18.31                          & 15.49               & 30.71               & 15.82                  & 33.94            & 50.22          & 83.93                       & 48.84                & 50.52             & 57.89            & 69.49            & 68.29               \\
OccNet &  0.50  & 47.29         & 59.06            & 20.63            & 48.32        & 63.05        & 24.12                          & 20.24               & 41.82               & 18.84                  & 23.38            & 41.12          & 86.46                       & 53.12                & 52.03             & 59.14            & 71.55            & 73.68               \\
\rowcolor[RGB]{220,220,220}OccNet & 0.25 & 53.00            & 65.93            & 22.84            & 64.09        & 72.69        & 32.73                          & 28.73               & 52.21               & 17.64                  & 22.05            & 51.26          & 89.05                       & 57.41                & 58.06             & 64.30            & 75.09            & 73.92               \\
\bottomrule
\end{tabular}%
}
\vspace{0.5mm}
\caption{\textbf{The performance of OccNet with ResNet50 backbone on nuScenes validation set for LiDAR segmentation task.} The method with the smallest $\Delta s$ show best performance.}
\label{tab:lidarseg}
\end{table*}

\begin{figure*}[tbh]
  \centering
    \includegraphics[width=0.9\linewidth]{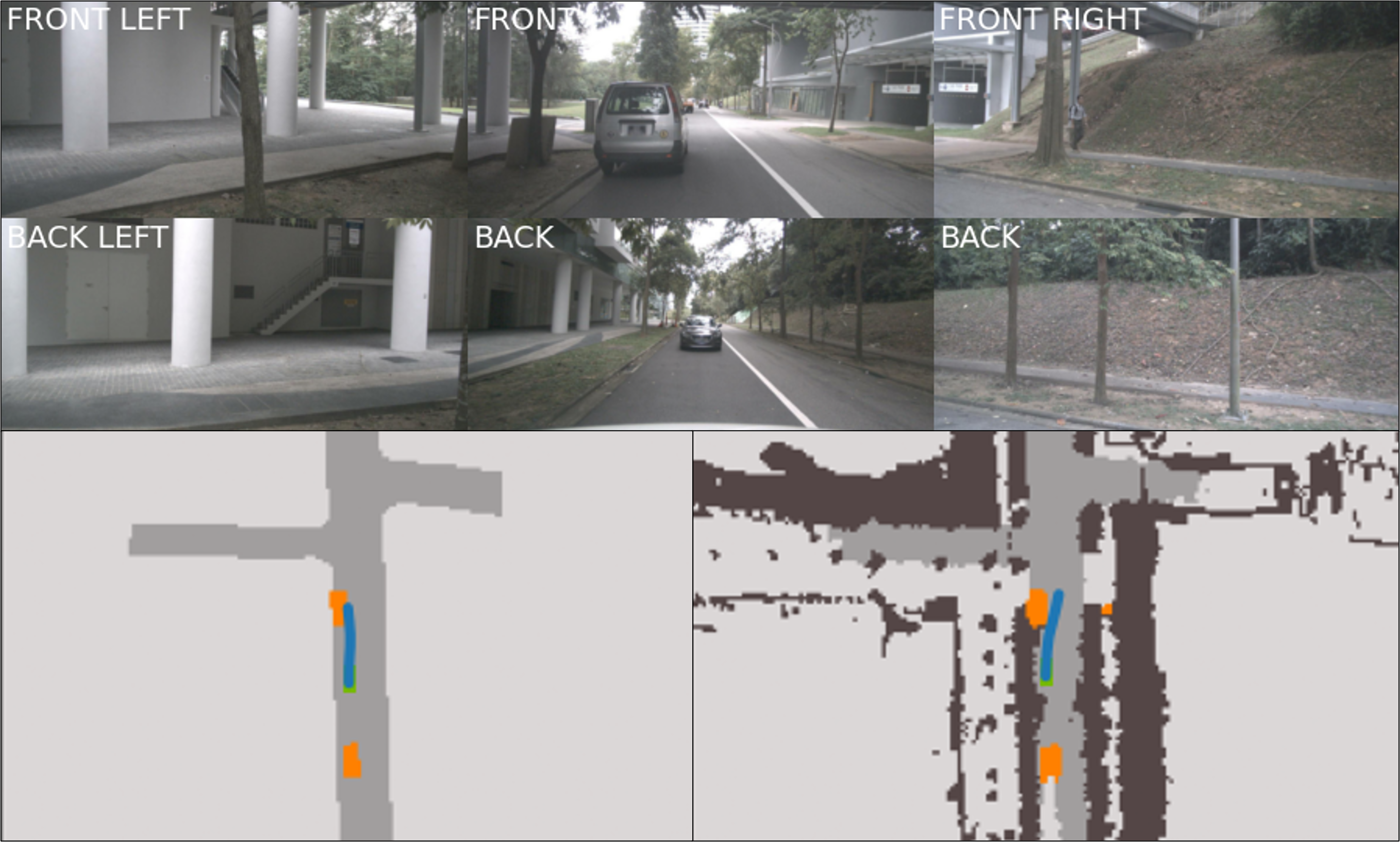}
  \caption{\textbf{Visualization of planning.} The blue line represents the planned trajectory, and the lower figures are rasterisation results of bounding box and occupancy, respectively. The trajectory obtained by the rasterized occupancy input can maintain a greater safety distance from the truck, because of the more accurate polygon representation.}
  \label{fig:planning2}
\end{figure*}

\section{Visualization Results}
\begin{figure*}[tbh]
  \centering
    \includegraphics[width=0.95 \linewidth]{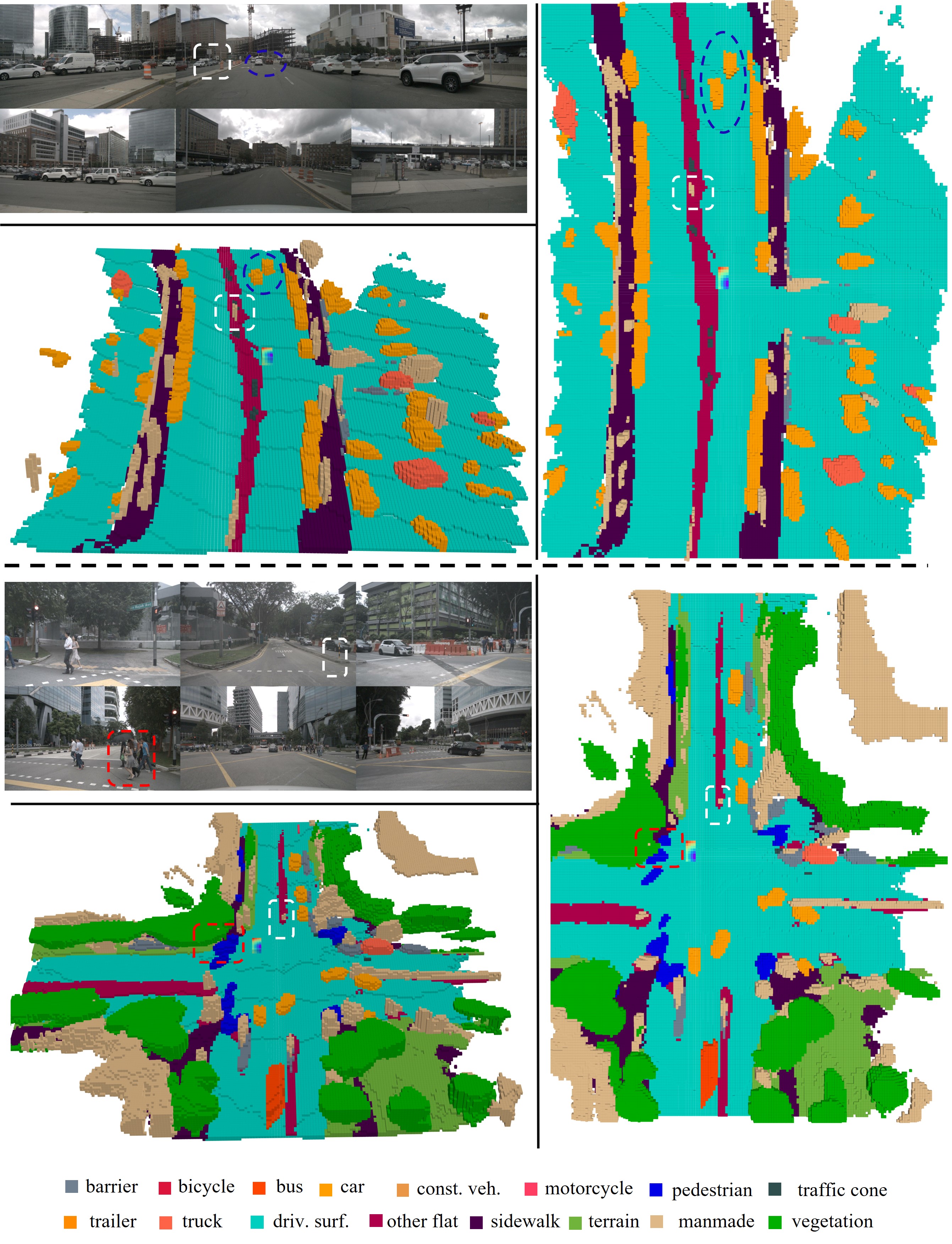}
  \caption{\textbf{Visualization of occupancy prediction.} For each scene, the top left figure is the surrounding camera input, and the left bottom figure and right figure represents the perspective view and top view of occupancy prediction result. The dashed region denotes that OccNet can predict the small size target or the distance target well.}
  \label{fig:occ_pred}
\end{figure*}

We sample two scenes in the validation set and provide detailed visualization of the occupancy prediction in Figure~\ref{fig:occ_pred}, indicating that OccNet can describe the scene geometry and semantics in detail.
As shown in Figure~\ref{fig:planning2}, we compare the rasterized occupancy with the rasterized bounding box as the input of planning module, indicating that occupancy is superior to bounding box for motion planning task.

%% file: arxiv.bbl
\begin{thebibliography}{10}\itemsep=-1pt

\bibitem{mobile2020ces}
{CES 2020 by Mobileye }.
\newblock \url{https://youtu.be/HPWGFzqd7pI}, 2020.

\bibitem{tesla_ai_day}
{Tesla AI Day}.
\newblock \url{https://www.youtube.com/watch?v=j0z4FweCy4M}, 2021.

\bibitem{behley2019semantickitti}
Jens Behley, Martin Garbade, Andres Milioto, Jan Quenzel, Sven Behnke, Cyrill
  Stachniss, and Jurgen Gall.
\newblock Semantickitti: A dataset for semantic scene understanding of lidar
  sequences.
\newblock In {\em ICCV}, pages 9297--9307, 2019.

\bibitem{caesar2020nuscenes}
Holger Caesar, Varun Bankiti, Alex~H Lang, Sourabh Vora, Venice~Erin Liong,
  Qiang Xu, Anush Krishnan, Yu Pan, Giancarlo Baldan, and Oscar Beijbom.
\newblock nuscenes: A multimodal dataset for autonomous driving.
\newblock In {\em CVPR}, pages 11621--11631, 2020.

\bibitem{cao2022monoscene}
Anh-Quan Cao and Raoul de Charette.
\newblock Monoscene: Monocular 3d semantic scene completion.
\newblock In {\em CVPR}, pages 3991--4001, 2022.

\bibitem{cao2022scenerf}
Anh-Quan Cao and Raoul de Charette.
\newblock Scenerf: Self-supervised monocular 3d scene reconstruction with
  radiance fields.
\newblock {\em arXiv preprint arXiv:2212.02501}, 2022.

\bibitem{chen2022persformer}
Li Chen, Chonghao Sima, Yang Li, Zehan Zheng, Jiajie Xu, Xiangwei Geng,
  Hongyang Li, Conghui He, Jianping Shi, Yu Qiao, et~al.
\newblock Persformer: 3d lane detection via perspective transformer and the
  openlane benchmark.
\newblock In {\em ECCV}, pages 550--567. Springer, 2022.

\bibitem{deng2009imagenet}
Jia Deng, Wei Dong, Richard Socher, Li-Jia Li, Kai Li, and Li Fei-Fei.
\newblock Imagenet: A large-scale hierarchical image database.
\newblock In {\em CVPR}, pages 248--255, 2009.

\bibitem{Fang2023}
Ming Fang and Zhiqi Li.
\newblock occupancy-for-nuscenes.
\newblock \url{https://github.com/FANG-MING/occupancy-for-nuscenes}, 2023.

\bibitem{fong2022panoptic}
Whye~Kit Fong, Rohit Mohan, Juana~Valeria Hurtado, Lubing Zhou, Holger Caesar,
  Oscar Beijbom, and Abhinav Valada.
\newblock Panoptic nuscenes: A large-scale benchmark for lidar panoptic
  segmentation and tracking.
\newblock In {\em ICRA}, 2022.

\bibitem{Hartley2004}
R.~I. Hartley and A. Zisserman.
\newblock {\em Multiple View Geometry in Computer Vision}.
\newblock Cambridge University Press, ISBN: 0521540518, second edition, 2004.

\bibitem{he2016resnet}
Kaiming He, Xiangyu Zhang, Shaoqing Ren, and Jian Sun.
\newblock Deep residual learning for image recognition.
\newblock In {\em CVPR}, pages 770--778, 2016.

\bibitem{st-p32022}
Shengchao Hu, Li Chen, Penghao Wu, Hongyang Li, Junchi Yan, and Dacheng Tao.
\newblock St-p3: End-to-end vision-based autonomous driving via
  spatial-temporal feature learning.
\newblock In {\em ECCV}, pages 533--549. Springer, 2022.

\bibitem{yang2022goal}
Yihan Hu, Jiazhi Yang, Li Chen, Keyu Li, Chonghao Sima, Xizhou Zhu, Siqi Chai,
  Senyao Du, Tianwei Lin, Wenhai Wang, Lewei Lu, Xiaosong Jia, Qiang Liu,
  Jifeng Dai, Yu Qiao, and Hongyang Li.
\newblock Planning-oriented autonomous driving.
\newblock In {\em CVPR}, 2023.

\bibitem{huang2022bevdet4d}
Junjie Huang and Guan Huang.
\newblock Bevdet4d: Exploit temporal cues in multi-camera 3d object detection.
\newblock {\em arXiv preprint arXiv:2203.17054}, 2022.

\bibitem{huang2021bevdet}
Junjie Huang, Guan Huang, Zheng Zhu, and Dalong Du.
\newblock Bevdet: High-performance multi-camera 3d object detection in
  bird-eye-view.
\newblock {\em arXiv preprint arXiv:2112.11790}, 2021.

\bibitem{huang2023tri}
Yuanhui Huang, Wenzhao Zheng, Yunpeng Zhang, Jie Zhou, and Jiwen Lu.
\newblock Tri-perspective view for vision-based 3d semantic occupancy
  prediction.
\newblock {\em arXiv preprint arXiv:2302.07817}, 2023.

\bibitem{li2021hdmapnet}
Qi Li, Yue Wang, Yilun Wang, and Hang Zhao.
\newblock Hdmapnet: An online hd map construction and evaluation framework.
\newblock In {\em ICRA}, pages 4628--4634. IEEE, 2022.

\bibitem{li2022bevdepth}
Yinhao Li, Zheng Ge, Guanyi Yu, Jinrong Yang, Zengran Wang, Yukang Shi,
  Jianjian Sun, and Zeming Li.
\newblock Bevdepth: Acquisition of reliable depth for multi-view 3d object
  detection.
\newblock {\em arXiv preprint arXiv:2206.10092}, 2022.

\bibitem{li2023voxformer}
Yiming Li, Zhiding Yu, Christopher Choy, Chaowei Xiao, Jose~M Alvarez, Sanja
  Fidler, Chen Feng, and Anima Anandkumar.
\newblock Voxformer: Sparse voxel transformer for camera-based 3d semantic
  scene completion.
\newblock {\em arXiv preprint arXiv:2302.12251}, 2023.

\bibitem{li2022bevformer}
Zhiqi Li, Wenhai Wang, Hongyang Li, Enze Xie, Chonghao Sima, Tong Lu, Yu Qiao,
  and Jifeng Dai.
\newblock Bevformer: Learning bird’s-eye-view representation from
  multi-camera images via spatiotemporal transformers.
\newblock In {\em ECCV}, pages 1--18. Springer, 2022.

\bibitem{liang2022bevfusion}
Tingting Liang, Hongwei Xie, Kaicheng Yu, Zhongyu Xia, Zhiwei Lin, Yongtao
  Wang, Tao Tang, Bing Wang, and Zhi Tang.
\newblock Bevfusion: A simple and robust lidar-camera fusion framework.
\newblock {\em arXiv preprint arXiv:2205.13790}, 2022.

\bibitem{lin2017feature}
Tsung-Yi Lin, Piotr Doll{\'a}r, Ross Girshick, Kaiming He, Bharath Hariharan,
  and Serge Belongie.
\newblock Feature pyramid networks for object detection.
\newblock In {\em CVPR}, pages 2117--2125, 2017.

\bibitem{Lin2017FocalLF}
Tsung-Yi Lin, Priya Goyal, Ross Girshick, Kaiming He, and Piotr Doll{\'a}r.
\newblock Focal loss for dense object detection.
\newblock In {\em ICCV}, pages 2980--2988, 2017.

\bibitem{liu2022bevfusion}
Zhijian Liu, Haotian Tang, Alexander Amini, Xingyu Yang, Huizi Mao, Daniela
  Rus, and Song Han.
\newblock Bevfusion: Multi-task multi-sensor fusion with unified bird's-eye
  view representation.
\newblock {\em arXiv preprint arXiv:2205.13542}, 2022.

\bibitem{loshchilov2017decoupled}
Ilya Loshchilov and Frank Hutter.
\newblock Decoupled weight decay regularization.
\newblock {\em arXiv preprint arXiv:1711.05101}, 2017.

\bibitem{miao2023occdepth}
Ruihang Miao, Weizhou Liu, Mingrui Chen, Zheng Gong, Weixin Xu, Chen Hu, and
  Shuchang Zhou.
\newblock Occdepth: A depth-aware method for 3d semantic scene completion.
\newblock {\em arXiv:2302.13540}, 2023.

\bibitem{mildenhall2020nerf}
Ben Mildenhall, Pratul~P. Srinivasan, Matthew Tancik, Jonathan~T. Barron, Ravi
  Ramamoorthi, and Ren Ng.
\newblock Nerf: Representing scenes as neural radiance fields for view
  synthesis.
\newblock In {\em ECCV}, 2020.

\bibitem{milioto2019rangenet++}
Andres Milioto, Ignacio Vizzo, Jens Behley, and Cyrill Stachniss.
\newblock Rangenet++: Fast and accurate lidar semantic segmentation.
\newblock In {\em IROS}, pages 4213--4220. IEEE, 2019.

\bibitem{moras2015grid}
Julien Moras, J. Dezert, and Benjamin Pannetier.
\newblock Grid occupancy estimation for environment perception based on belief
  functions and pcr6.
\newblock volume 9474, 04 2015.

\bibitem{mueller2022instant}
Thomas M\"uller, Alex Evans, Christoph Schied, and Alexander Keller.
\newblock Instant neural graphics primitives with a multiresolution hash
  encoding.
\newblock {\em ACM Trans. Graph.}, 41(4):102:1--102:15, July 2022.

\bibitem{Silberman2012nyuv2}
Pushmeet~Kohli Nathan~Silberman, Derek~Hoiem and Rob Fergus.
\newblock Indoor segmentation and support inference from rgbd images.
\newblock In {\em ECCV}, 2012.

\bibitem{Shi2020PVRCNNPF}
Shaoshuai Shi, Chaoxu Guo, Li Jiang, Zhe Wang, Jianping Shi, Xiaogang Wang, and
  Hongsheng Li.
\newblock Pv-rcnn: Point-voxel feature set abstraction for 3d object detection.
\newblock In {\em CVPR}, pages 10529--10538, 2020.

\bibitem{song2016ssc}
Shuran Song, Fisher Yu, Andy Zeng, Angel~X Chang, Manolis Savva, and Thomas
  Funkhouser.
\newblock Semantic scene completion from a single depth image.
\newblock In {\em CVPR}, pages 1746--1754, 2017.

\bibitem{Tancik2022blocknerf}
Matthew Tancik, Vincent Casser, Xinchen Yan, Sabeek Pradhan, Ben Mildenhall,
  Pratul~P. Srinivasan, Jonathan~T. Barron, and Henrik Kretzschmar.
\newblock Block-nerf: Scalable large scene neural view synthesis.
\newblock In {\em CVPR}, pages 8248--8258, June 2022.

\bibitem{wang2022monocular}
Tai Wang, Jiangmiao Pang, and Dahua Lin.
\newblock Monocular 3d object detection with depth from motion.
\newblock In {\em European Conference on Computer Vision}, pages 386--403.
  Springer, 2022.

\bibitem{wang2021fcos3d}
Tai Wang, Xinge Zhu, Jiangmiao Pang, and Dahua Lin.
\newblock Fcos3d: Fully convolutional one-stage monocular 3d object detection.
\newblock In {\em ICCV}, pages 913--922, 2021.

\bibitem{wang2023openoccupancy}
Xiaofeng Wang, Zheng Zhu, Wenbo Xu, Yunpeng Zhang, Yi Wei, Xu Chi, Yun Ye,
  Dalong Du, Jiwen Lu, and Xingang Wang.
\newblock Openoccupancy: A large scale benchmark for surrounding semantic
  occupancy perception.
\newblock {\em arXiv preprint arXiv:2303.03991}, 2023.

\bibitem{zhou2022cross}
Brady Zhou and Philipp Kr{\"a}henb{\"u}hl.
\newblock Cross-view transformers for real-time map-view semantic segmentation.
\newblock In {\em CVPR}, pages 13760--13769, 2022.

\bibitem{Zhu2021DeformableDD}
Xizhou Zhu, Weijie Su, Lewei Lu, Bin Li, Xiaogang Wang, and Jifeng Dai.
\newblock {Deformable DETR}: Deformable transformers for end-to-end object
  detection.
\newblock In {\em ICLR}, 2021.

\bibitem{zhu2021cylindrical}
Xinge Zhu, Hui Zhou, Tai Wang, Fangzhou Hong, Yuexin Ma, Wei Li, Hongsheng Li,
  and Dahua Lin.
\newblock Cylindrical and asymmetrical 3d convolution networks for lidar
  segmentation.
\newblock In {\em CVPR}, pages 9939--9948, 2021.

\end{thebibliography}
